%% file: main_R2.tex
\documentclass[10pt,twocolumn,letterpaper]{article}

\usepackage{wacv}              
\usepackage{algorithm}
\usepackage{multirow}
\usepackage{tikz}
\usepackage{tcolorbox}
\usepackage{xcolor}
\usepackage{algpseudocode}
\usepackage[table]{xcolor}
\definecolor{lightred}{rgb}{1,0.8,0.8}   
\definecolor{lightyellow}{rgb}{1,1,0.6}  
\definecolor{lightgreen}{rgb}{0.8,1,0.8} 

\newcommand{\lightgreenhl}[1]{{\sethlcolor{lightgreen}\hl{#1}}}
\definecolor{lightblue}{rgb}{0.7,0.85,1} 
\newcommand{\lightbluehl}[1]{{\sethlcolor{lightblue}\hl{#1}}}

\usepackage{colortbl}
\usepackage{array}
\usepackage{soul}
\usepackage{booktabs}
\usepackage{soul}        

\definecolor{wacvblue}{rgb}{0.21,0.49,0.74}
\usepackage[pagebackref,breaklinks,colorlinks,allcolors=wacvblue]{hyperref}

\usepackage{subdef}

\def\wacvPaperID{1485} 
\def\confName{WACV}
\def\confYear{2026}

\title{Tables Decoded: \name{} for Structure, \tabqa{} for Understanding}

\author{
Jahanvi Rajput$^{*}$$^{1}$,
Dhruv Kudale\thanks{Authors contributing equally.} $^{1}$\thanks{Work done while pursuing MS at IIT Bombay.},
Saikiran Kasturi$^{1}$,
Utkarsh Verma$^{1}$,
Ganesh Ramakrishnan$^{1,2}$ \\
{\tt\small \{23d0378, 22m2116, 24m2157, 24m2153, ganramkr\}@iitb.ac.in} \\[4pt]
$^{1}$Indian Institute of Technology Bombay, $^{2}$BharatGen
}

\begin{document}
\maketitle
\input{sec_R2/0_abstract}    
\input{sec_R2/1_intro}
\input{sec_R2/2_related_work}
\input{sec_R2/3_tab_rec}

\input{sec_R2/4_exp_results}

\input{sec_R2/5_conclusion}

\input{sec_R2/6_acknowledgement}

{
    \small
    \bibliographystyle{ieeenat_fullname}
    \bibliography{main}
}


\input{sec_R2/supplementary}


\end{document}

%% file: sec_R2/0_abstract.tex
\begin{abstract}

Table understanding is a core task in document intelligence, encompassing two key subtasks: table reconstruction and table visual question answering (TabVQA). While recent approaches predominantly rely on vision-language models (VLMs) operating on table images, we propose a more scalable and effective alternative based on structured textual representations. These representations are easier to process, align more naturally with LLMs, and eliminate the need for language-specific visual encoders, making them particularly suitable for multilingual documents. We present \name{}, which separates physical structure recognition, logical structure recognition, and OCR to extract both layout and content accurately. \name{} outputs tables in Optimised Table Structure Language (OTSL), a compact and unified format that encodes cell arrangements and textual content. On table structure recognition (TSR), \name{} achieves TEDS-Structure scores comparable with state-of-the-art methods across FinTabNet, PubTabNet, and PubTables-1M. We further establish its robustness on non-English tables through our curated Hindi benchmark, \bench{}. Building on this, we introduce \tabqa{}, an LLM fine-tuned on OTSL sequences. Our approach yields gains of 9.3 p.p. on WTQ (TabQA) and 9.2 p.p. on FinTabNetQA (TabVQA), respectively. On \bench{}, our method ranks second among all VLMs and \name{} + LLM variants. We release our code, models, and benchmark at: \url{https://github.com/Tihiitborg/Tables-Decoded}.



\end{abstract}

%% file: sec_R2/1_intro.tex
\section{Introduction}
\label{sec:intro}

\begin{figure*}
    \centering
    \fbox{\includegraphics[width=0.85\linewidth]{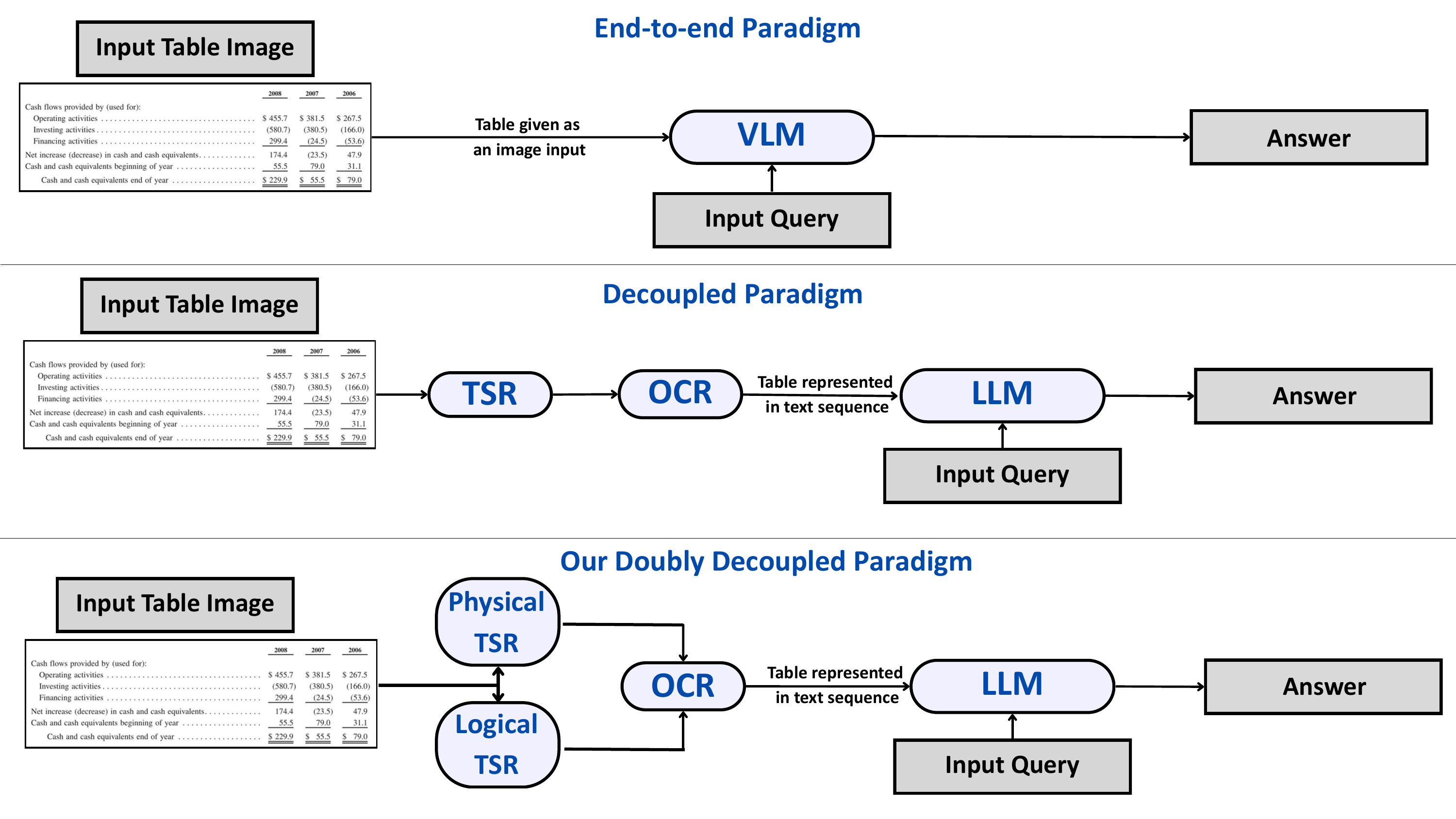}}
    \caption{TabVQA paradigms: (i) end-to-end, where a VLM directly processes the table image and question; (ii) decoupled, where TSR + OCR extracts structure and content into text (HTML, OTSL) for an LLM; and (iii) our doubly decoupled approach, which further separates physical and logical TSR for better control. The steps following OCR remain the same as for the conventional decoupled approach.}
    \label{fig:teaser}
\end{figure*}

Tables are a vital medium for structured information in domains like science, finance, and administration. Their visual layouts encode relational data, supporting tasks such as retrieval, parsing, summarization, and Table Visual Question Answering (TabVQA). Real-world tables, however, present challenges: diverse layouts, formatting variations, multiple languages, and complications like merged cells, spanning, or noisy scans. Broadly, as seen in Figure \eqref{fig:teaser}, existing approaches to table understanding follow one of the many distinct paradigms. The first relies on Vision Language Models (VLMs), which directly consume table images and jointly model visual and textual features using architectures pretrained on large multimodal datasets. These models are typically fine-tuned end-to-end for tasks such as table detection, reconstruction, or TabVQA. The second paradigm leverages Large Language Models (LLMs), which operate entirely in the language space. In this setting, tables are first serialized into textual formats, such as HTML, Markdown, or plain-text formats, before being provided as input prompts to LLMs (Figure \eqref{fig:teaser}). These models then reason over these serialized tables to answer queries. While VLMs benefit from strong visual grounding, they are often monolithic and inflexible, tightly coupling layout reasoning, content extraction, and question answering into a single opaque pipeline. This end-to-end dependency makes them difficult to debug or adapt. Moreover, they require extensive pretraining with image-text pairs. VLMs also show a performance gap, especially for non-Latin scripts or low-resource languages. Crucially, the internal representations they learn are often not interpretable or easily transferable to new tasks. In contrast, LLM-based approaches offer a modular, language-agnostic alternative. Once tables are converted into structured textual formats, general-purpose LLMs can perform downstream tasks like TabVQA with minimal fine-tuning. However, the challenge lies in the table serialization process itself. An important factor that significantly impacts table understanding performance, especially when leveraging LLMs, is the choice of table representation. Simple plain-text formats fail to preserve structural information such as rows, columns, and cells, making it difficult for models to infer relationships between table elements. While formats like LaTeX and HTML can accurately capture table structures, they are often verbose and lead to long token sequences, which strain the input limitations of most LLMs and reduce overall efficiency. Markdown, though more compact, cannot faithfully represent complex structures involving merged cells, making it unsuitable for complex tables. These limitations highlight the need for a compact format, one that preserves both structure and content while remaining well-suited for LLM-based processing. To address this, we advocate the use of the recently proposed Optimised Table Structure Language (OTSL) sequences \cite{otsl}. OTSL provides a compact yet expressive linear representation of tables, capturing hierarchical layouts, spanning cells, and textual content in a form shorter than HTML while preserving structure. We design a modular pipeline that is doubly decoupled (Figure \eqref{fig:teaser}): first, by separating Table Structure Recognition (TSR) and Optical Character Recognition (OCR) to handle structure and text independently, and second, within TSR itself by disentangling physical (rows, columns) from logical (cell arrangement) structure. This design improves interpretability, robustness across document styles and languages, and scalability for TabVQA. It outperforms recent end-to-end VLMs on standard benchmarks and enables multilingual TabVQA, including low-resource scripts, by shifting reasoning to text-based LLMs using OTSL. Our main contributions are as follows:

\begin{itemize}
    \item We present \name, a \textbf{D}oubly d\textbf{E}coupled tab\textbf{L}e recons\textbf{T}ruction \textbf{A}pproach that doubly decouples TSR followed by OCR to produce a compact OTSL sequence with high structural fidelity, outperforming recent VLMs.
    
    \item To support downstream tasks, we introduce a lossless algorithm to convert widely used HTML format into OTSL, effectively capturing both table structure and content. 
    
    \item We propose \tabqa{}, \textbf{TA}ble structu\textbf{R}e-aware \textbf{Q}uestion \textbf{A}nswering, an LLM fine-tuned on OTSL-formatted tables for TabVQA, demonstrating stronger performance over models fine-tuned on HTML and other baselines on WikiTableQuestions  (WTQ) and FinTabNetQA datasets.

    \item Finally, we introduce and release \bench{}, \textbf{T}able \textbf{O}riented \textbf{R}econstruction and \textbf{Q}uestion-answering \textbf{U}pon d\textbf{E}vanagari, a new Hindi benchmark to demonstrate the multilingual capability of our framework, showcasing effective performance on both table reconstruction and TabVQA tasks for the Hindi language.

\end{itemize}


%% file: sec_R2/2_related_work.tex
\section{Related Literature}
\label{sec:Related work}

\textbf{End-to-End Table Reconstruction} leverages VLMs for holistic table understanding, aiming to directly generate structured representations (HTML or JSON) from table images. These models use pre-trained vision-language encoders to jointly process visual and textual cues, enabling the prediction of table structures and content. Approaches like MTLTabNet \cite{mtl-tab-net}, SmolDocLing \cite{nassar2025smoldocling}, SmolVLM \cite{smolvlm}, and Granite-Vision \cite{granitevision} fall into this category. The key advantage of this paradigm is its ability to unify structure and content prediction without relying on intermediate OCR outputs or handcrafted rules. However, these models often struggle with multilingual scripts, complex layouts, and noisy scanned documents, particularly when such variations are underrepresented in training data.

\textbf{Decoupled Table Reconstruction} involves two key steps: TSR and OCR. TSR models are responsible for identifying cell boundaries and relationships. Object detection-based TSR methods focus on reconstructing the physical layout of tables by localizing cells using models like Faster R-CNN \cite{faster-rcnn}, Mask R-CNN \cite{mask-rcnn}, YOLO \cite{yolo}, etc. Transformer-based variants such as DETR \cite{detr}, TATR \cite{tatr-pub-1m}, TableFormer \cite{tableformer}, and TSRFormer \cite{tsrformer3} have also been equipped for this task. However, such object detection-based TSR often depends on post-processing for mapping cells to row and column numbers, making them sensitive to detection errors. Later, Im2Seq-based TSR methods dominated the field of TSR, where they directly predict the logical structures (cell mapping) from images using encoder-decoder architectures, outputting formats like HTML or LATEX. Recent approaches adopt compact representations like OTSL \cite{otsl, kudale2025sprint} to improve inference efficiency and structural consistency. Hybrid TSR methods, including graph-based models like GTE \cite{gte} and TGRNet \cite{tgrnet}, treat table reconstruction as a graph problem. Others combine visual detection with token-level generation, such as EDD \cite{zhong2020image}, local attention-based TSR \cite{tsr-local-attention}, or use visual and positional cues to refine predictions \cite{lgpma, tabstruct, vast}. The emergence of Im2Seq models has often offered better end-to-end consistency by synchronising physical and logical structure predictions in a unified pipeline. Table reconstruction datasets such as PubTabNet \cite{zhong2020image}, FinTabNet \cite{fintabnet}, PubTables \cite{tatr-pub-1m}, SynthTabNet \cite{synthtabnet}, and TabRecSet \cite{tabrecset} provide diverse annotations, with evaluation metrics like TEDS \cite{zhong2020image} and GriTS \cite{grits} assessing both spatial and structural accuracy. Once the structure is extracted, OCR engines are used to recognize the textual content within each cell. Popular OCR engines suitable for this task include Tesseract \cite{tesseract}, DocTR \cite{doctr}, EasyOCR \cite{easyocr}, and several others \cite{surya, paddleocr}. 

\textbf{Table-based Question Answering} (TabQA) focuses on reasoning over structured table data, typically provided in CSV or HTML format, to answer natural language questions. Recently, LLMs \cite{grattafiori2024llama, jiang2024mixtral, qwen} have been fine-tuned or prompted for direct reasoning over tabular inputs, exhibiting strong zero-shot and few-shot capabilities. However, their performance heavily depends on the quality and format of both the table and the prompt. Thus, the choice of representation used to encode tabular data becomes a critical design decision in decoupled \textbf{TabVQA} pipelines. TabVQA, in contrast, integrates visual understanding with question answering, operating directly on table images. Several datasets, such as ComTQA \cite{tabpedia}, WTQ \cite{wtq}, and FinTabNetQA \cite{kim2024tablevqa_bench}, have been introduced to support this task. As discussed earlier, there are two main approaches to TabVQA \cite{kim2024tablevqa_bench}: (i) VLM-based models \cite{hu2024mplug_docowl, llava, blip2, qwen, cogagent, wang2023cogvlm} that jointly encode visual and textual information to generate answers, and (ii) decoupled pipelines that first extract table structure and content via TSR + OCR, then apply LLMs for textual question answering. We build upon this paradigm, where \name{} handles the doubly decoupled table reconstruction, followed by \tabqa{}, our finetuned LLM, for question answering. Both components of our approach are detailed in the upcoming  Section \eqref{sec: Table Reconstruction}.

%% file: sec_R2/3_tab_rec.tex
\section{Our Methodology}
\label{sec: Table Reconstruction}
Our methodology has two components: \name{} for generating OTSL from table images, and \tabqa{} for TabVQA on these sequences extracted from input table images. 

\begin{figure*}[]
    \centering
    \fbox{\includegraphics[width=0.85\textwidth]{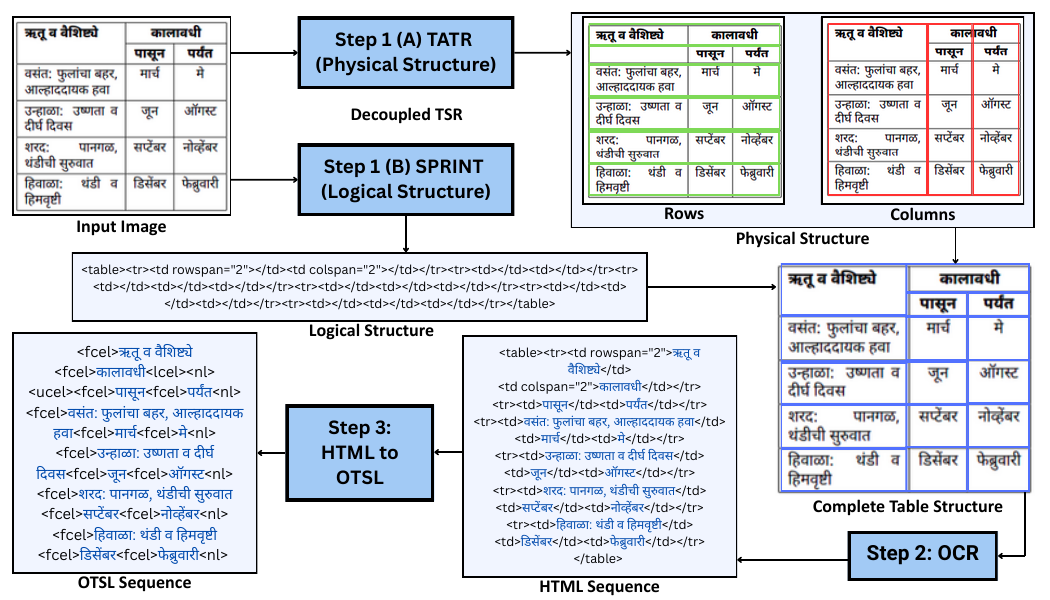}}
    \caption{\name{} overview: given a table image, the system first performs decoupled TSR (physical structure through TATR and logical structure through SPRINT), followed by OCR to extract table content. The resulting HTML is converted into a compact OTSL sequence.}
    \label{fig:delta-pipeline}
\end{figure*}

\subsection{\textbf{\name{}} for Structure}

Figure \eqref{fig:delta-pipeline} presents an overview of \name{}. Given a table image, we perform TSR followed by OCR to extract the table content, where TSR is also divided into two components: physical structure and logical structure. The final output after TSR and OCR is an HTML representation, which is then transformed into an OTSL sequence in a lossless manner using Algorithm \eqref{algo_html_to_otsl}.  Overall, our methodology for \name{} consists of the following key steps:


\subsubsection{TSR for Cell Demarcation}

For TSR, we employ a combination of SPRINT \cite{kudale2025sprint} and TATR \cite{smock2023aligning}, which together provide a comprehensive understanding of table layouts. We chose SPRINT as it has shown fast, robust, and language-agnostic performance on logical TSR, making it ideal for capturing cell relationships across diverse scripts. SPRINT provides an HTML sequence corresponding to the input table's logical structure. TATR, also a recent state-of-the-art model, complements this by focusing on physical structure recognition, accurately identifying rows and columns to demarcate individual cells in the table image. The combined output is a structured HTML tag sequence that represents the table, including complex features such as cell merges, row spans, and column spans, with each \texttt{<td>} tag attributed with precise bounding box coordinates. The output of the TSR step with cells highlighted is seen in Figure \eqref{fig:delta-pipeline}. We provide a more detailed explanation of how TATR and SPRINT work together to generate the final HTML sequence in the supplementary material.

\subsubsection{OCR for Content Extraction}
The TSR output, an HTML tag sequence with bounding box annotations, is used to perform OCR on each cell. We employ EasyOCR ~\cite{easyocr} for its seamless integration with TSR, GPU compatibility, and fast processing, while remaining modular and easily replaceable. For each \texttt{<td>} tag, the bounding box crops the corresponding cell from the table image, and EasyOCR extracts the text, which is then embedded back into the HTML string. This enriches the structure with content and extends naturally to multilingual tables. The choice of EasyOCR is further validated through an ablation study with other OCR models, detailed in the supplementary material. 



\subsubsection{HTML to OTSL conversion}
\begin{figure}[h]
\centering
\scriptsize

\begin{minipage}[t]{0.33\textwidth}
\centering
\begin{tikzpicture}[scale=0.8]
\draw (0,0) rectangle (3,2);
\draw (0,1) -- (3,1);
\draw (1,1) -- (1,0);
\draw (2,1) -- (2,0);
\draw (1,0.5) -- (3,0.5);

\node at (1.5,1.5) {\textbf{ABC}};
\node at (0.5,0.75) {\textbf{D}};
\node at (1.5,0.75) {\textbf{E}};
\node at (2.5,0.75) {\textbf{F}};
\node at (2.5,0.25) {\textbf{H}};
\end{tikzpicture}
\end{minipage}

\vspace{1mm}
$\downarrow$ {\scriptsize Using Table Reconstruction}
\vspace{1mm}

\begin{minipage}[t]{0.45\textwidth}
\centering
\begin{tcolorbox}[colback=white, colframe=black, boxrule=0.2mm, left=1pt,right=1pt,top=1pt,bottom=1pt]
 \centering \textbf{HTML} \\
\scriptsize
\begin{tt}
<html><table><tr><td colspan=3>\textcolor{blue}{ABC}</td></tr><tr><td rowspan=2>\textcolor{blue}{D}</td><td>\textcolor{blue}{E}</td><td>\textcolor{blue}{F}</td></tr><tr><td> </td><td>\textcolor{blue}{H}</td></tr></table></html>
\end{tt}
\end{tcolorbox}
\end{minipage}

\vspace{1mm}
$\downarrow$ {\scriptsize Using Algorithm \eqref{algo_html_to_otsl}}
\vspace{1mm}

\begin{minipage}[t]{0.45\textwidth}
\centering
\begin{tcolorbox}[colback=green!10!white, colframe=black, boxrule=0.2mm, left=1pt,right=1pt,top=1pt,bottom=1pt]
\centering \textbf{\textcolor{green!50!black}{Our Minimized OTSL}}\\
\scriptsize
\begin{tt}
<otsl><fcel>\textcolor{blue}{ABC}<lcel><lcel><nl><fcel>\textcolor{blue}{D}<fcel>\textcolor{blue}{E}<fcel> \textcolor{blue}{F}<nl><ucel><ecel><fcel>\textcolor{blue}{H}<nl></otsl>
\end{tt}
\end{tcolorbox}
\end{minipage}

\caption{Illustration of converting a table grid into HTML and then into the OTSL sequence using Algorithm \eqref{algo_html_to_otsl}.}
\label{tab_rep}
\end{figure}
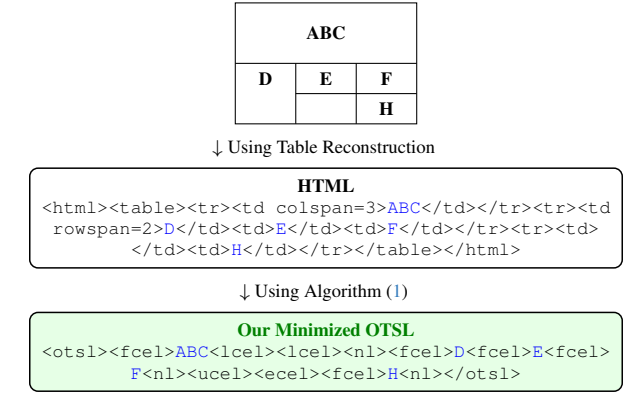

As highlighted in prior work \cite{otsl}, HTML representations are often large and noisy due to their verbose, repetitive tag structures, making them difficult to use directly with LLMs in TabQA tasks. To overcome these limitations, we convert the HTML output into an OTSL sequence, a compact, structured format that efficiently captures both the table’s layout and content. This conversion significantly reduces input length while preserving all necessary information, making it more suitable for LLM-based inference and improving performance. We present Algorithm \eqref{algo_html_to_otsl}, which outlines the process of converting an HTML sequence into its corresponding OTSL representation in a lossless manner. Figure \eqref{tab_rep} provides an example of a complex table, showcasing the HTML generated by \name, and the resulting OTSL sequence. This comparison demonstrates the compactness and structured clarity of OTSL, which enhances downstream task performance. We refer readers to the original OTSL specification \cite{otsl} for a detailed understanding of its syntax and token design.

\begin{algorithm}
\caption{Extract OTSL Matrix from HTML string}
\begin{algorithmic}[1]
\Require HTML string
\Ensure OTSL matrix string

\State Parse HTML, find \texttt{<table>}
\State Compute $R$ (rows) and $C$ (columns) using \texttt{rowspan}, \texttt{colspan}
\State Initialize $\text{otsl\_matrix}[R][C] \gets \texttt{"<ecel>"}$
\State Initialize $\text{cell\_map}[R][C] \gets 0$

\For{each row $i$}
    \State $col\_idx \gets 0$
    \For{each cell in row}
        \While{$\text{cell\_map}[i][col\_idx] = 1$} \State $col\_idx \gets col\_idx + 1$ \EndWhile
        \State Extract $rowspan$, $colspan$, text
        \State Assign \texttt{"<fcel>"} if text exists, else \texttt{"<ecel>"}
        \State Fill merged cells: \texttt{"<lcel>"} (left), \texttt{"<ucel>"} (up), \texttt{"<xcel>"} (cross)
        \State Update $\text{cell\_map}$, move to next column
    \EndFor
\EndFor

\State Convert $\text{otsl\_matrix}$ to string with \texttt{"<nl>"} separators
\State \Return OTSL matrix 

\end{algorithmic}
\label{algo_html_to_otsl}
\end{algorithm}

\begin{figure}[]
    \centering
    \fbox{\includegraphics[width=0.45\textwidth]{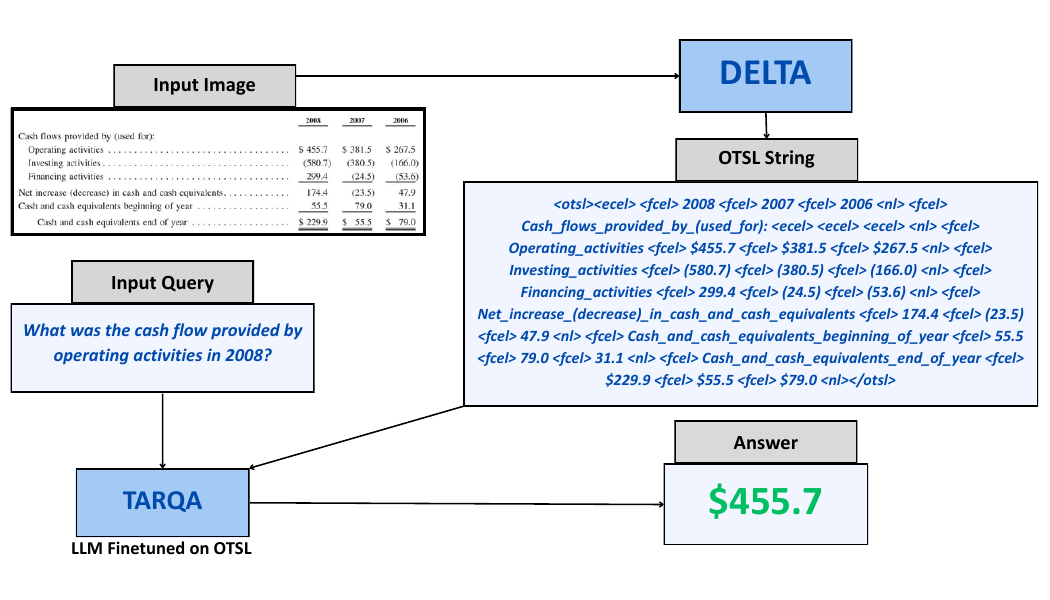}}
    \caption{\tabqa{}, an LLM fine-tuned on OTSL sequence, uses output from \name{} along with the query to generate answers.}
    \label{fig:tarka-pipeline}
\end{figure}

\subsection{\textbf{\tabqa{}} for Question Answering}
We introduce \tabqa{}, an LLM fine-tuned for TabQA directly on OTSL sequences. Its key contribution is to reframe table understanding as a text-to-text problem, leveraging OTSL, to eliminate the reliance on table images for downstream reasoning tasks. By operating purely on structured OTSL inputs, \tabqa{} naturally supports decoupled processing and multilingual extension.
\tabqa{} can be seamlessly paired with \name{}, which produces OTSL representations, thereby enabling a complete TabVQA pipeline as illustrated in Figure~\eqref{fig:tarka-pipeline}. The choice of OTSL over alternative formats, fine-tuning details, and other related experimentation is described in the upcoming Section \eqref{sec:experiments}.

%% file: sec_R2/4_exp_results.tex
\section{Experiments}
\label{sec:experiments}
Now that we have described \name{} for producing OTSL sequences, we proceed to validate the effectiveness of OTSL as the most suitable representation for TabQA (and eventually TabVQA). We conduct a series of experiments comparing OTSL with other common table representations. We first describe the datasets used, followed by a detailed explanation of the experimental setup. 



\subsection{Datasets}
We begin by describing the English and non-English table-based datasets we use in our experimental setup.
\subsubsection{English Datasets}
As \name{} is a framework composed of multiple off-the-shelf blocks, we skip training and directly infer and report results on FinTabNet (test), PubTables (test), and PubTabNet (validation, for fair comparison with prior work). For TabQA, we fine-tune LLMs on WTQ with different table representations (HTML, plain text, OTSL) and select the best-performing variant of \tabqa. Finally, we evaluate the full TabVQA pipeline of \name{} followed by \tabqa{} on FinTabNetQA. Dataset statistics are present in Table \eqref{data_tab}.

\begin{table}
\centering
\begin{tabular}{lccc}
\hline
\textbf{Dataset} & \textbf{Train} & \textbf{Val} & \textbf{Test} \\
\hline
FinTabNet   & -  & - & 10305 \\
PubTabNet   & - & 6942 & -        \\
PubTables   & - & -  & 92841 \\
WTQ & 11321 & - & 7175\\
FinTabNetQA & - & - & 250\\
\hline
\end{tabular}
\caption{Datasets used for reconstruction, TabQA and TabVQA}.
\label{data_tab}
\end{table}

\subsubsection{\bench Dataset}
\begin{figure}[h]
    \centering
    \begin{subfigure}{0.8\linewidth}
        \centering
        \includegraphics[width=\linewidth]{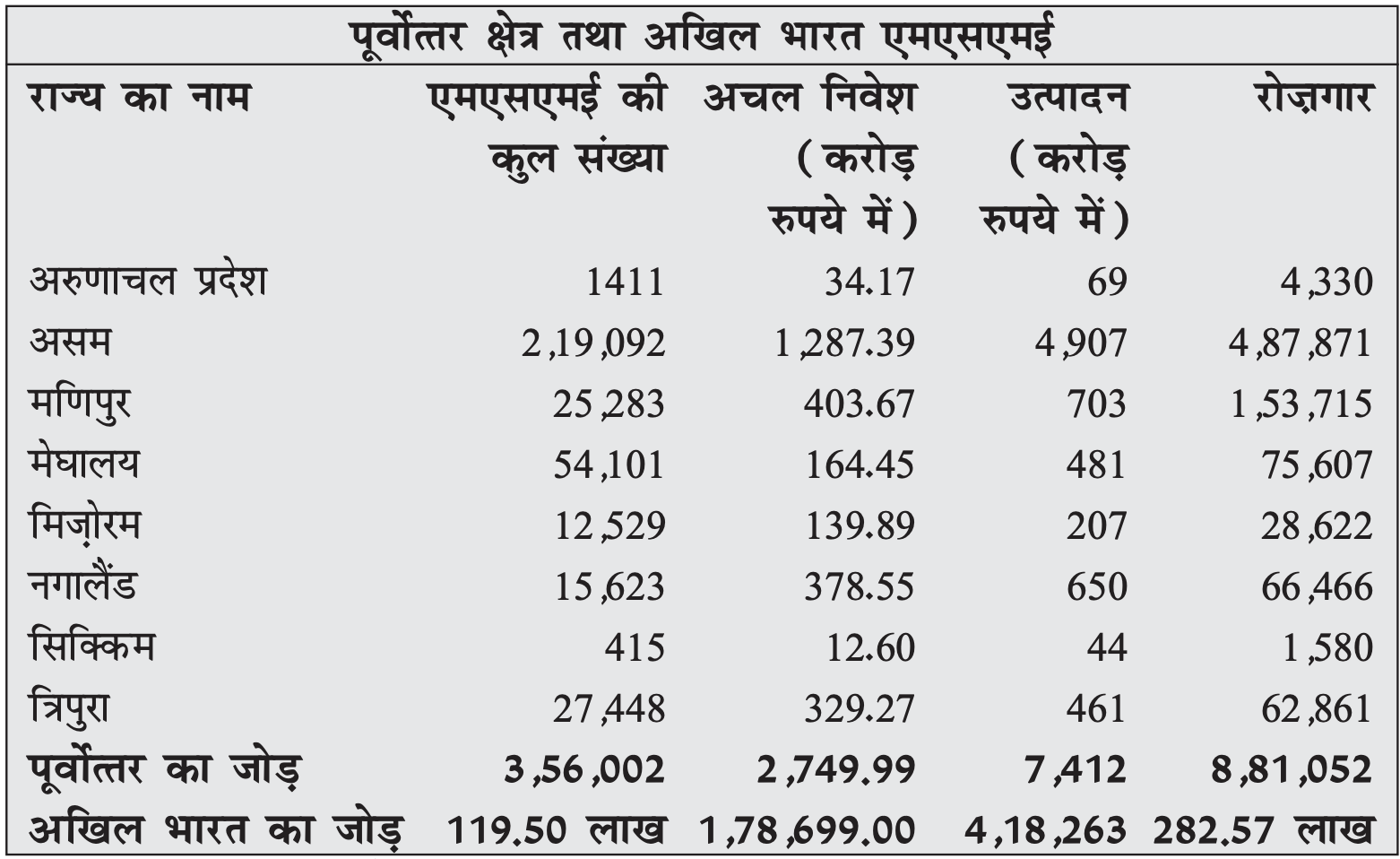}
    \end{subfigure}
    \hfill
    \begin{subfigure}{0.8\linewidth}
        \centering
\includegraphics[width=\linewidth]{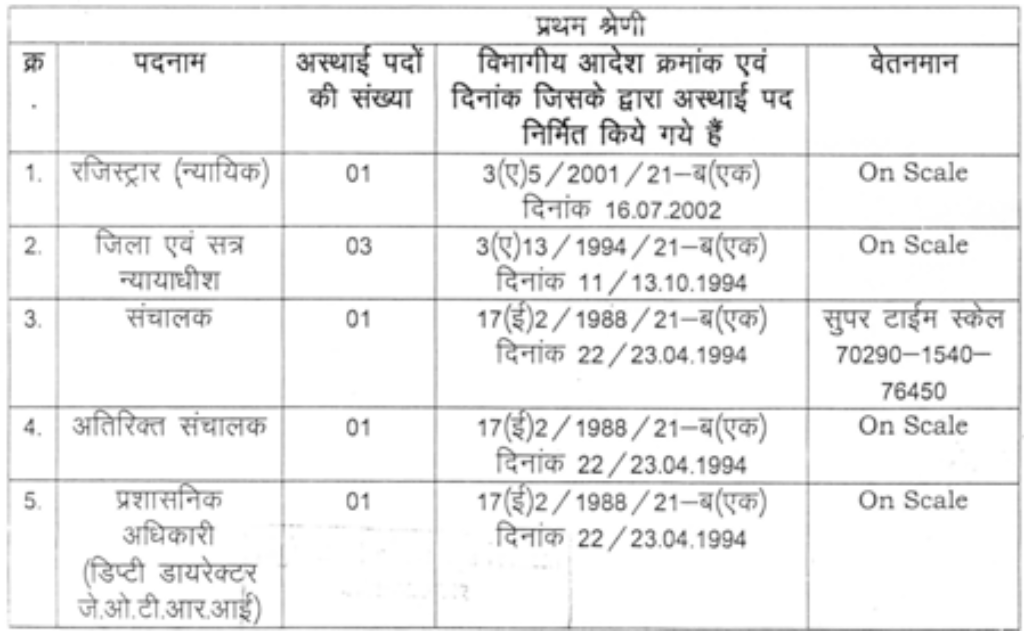}
    \end{subfigure}
    \caption{Sample Hindi images from the \bench{} dataset}
    \label{torque_sample}
\end{figure}

We introduce \bench
, a Hindi benchmark for evaluating both Hindi table reconstruction and Hindi TabVQA. It contains 210 tables : 109 scanned and 101 digital-born, cropped, and sourced from government circulars \cite{mphc, rajbhasha} and spiritual books from MUSTARD \cite{kudale2025sprint}, with a mix of simple (149) and complex (61) table structures. Figure \eqref{torque_sample} shows a few sample images from the dataset. The dataset also includes 422 manually verified QA pairs, generated using GPT-oss-20B \cite{agarwal2025gpt}. For table reconstruction, ChatGPT-4o \cite{openai2024gpt4technicalreport} outputs were manually post-corrected to obtain the exact ground-truth HTML sequences. \bench{} is used to benchmark open-source models and our proposed pipeline, showing that our approach is language-agnostic and outperforms conventional decoupled pipelines and significant VLMs without any Hindi-specific finetuning (zero-shot).

\subsection{Generating Table Representations}

The WTQ training set already contains HTML sequences (ground truth). To create inputs in other formats, we apply format-specific conversions: for OTSL, we use our proposed Algorithm \eqref{algo_html_to_otsl} to convert the HTML into a compact sequence; and for plain text, we parse the HTML content row-wise (left to right) and flatten it into a simple text string. Based on the chosen representation format, the corresponding LLM is fine-tuned for the TabQA task.

\subsection{Finetuning \textbf{\tabqa} Variants}
\label{sec:ftn}
We fine-tuned the Meta-LLaMA-3-8B-Instruct \cite{llama3modelcard} for the question-answering task to generate concise, accurate natural language answers based on a table OTSL sequence and a question. The model was fine-tuned using a custom instruction-tuned setup, where each example followed a structured prompt template that included an instruction, a serialised table, a natural language question, and the ground truth answer. Table~\eqref{fine-tuning} summarises the fine-tuning configuration of \tabqa{} on the WTQ dataset. The dataset was preprocessed using the HuggingFace AutoTokenizer with padding applied up to a maximum sequence length of 4096 tokens and truncation enabled. Model inference was carried out using greedy decoding. All experiments were conducted using a CUDA-enabled NVIDIA H100 80GB GPU device. The prompt used for fine-tuning is as follows:

\begin{small}
\begin{ttfamily}
\noindent
\#\#\# Prompt Used for Finetuning: \\
Given the following table, answer the question in one word or a short phrase. Do not provide an explanation.\\[1ex]
\#\#\# Table: \{OTSL\_sequence\}\\[1ex]
\#\#\# Question: \{User\_query\}\\[1ex]
\#\#\# Answer: \{answer\} \\
\end{ttfamily}
\end{small}

\begin{table}
\centering
\begin{tabular}{ll}
\toprule
\textbf{Parameter}            & \textbf{Value} \\ 
\midrule
Model                         & LLaMA-3-8B-Instruct \\
Optimizer                     & AdamW \\
Learning Rate                 & $2e^{-5}$ \\
Batch Size                    & 1 (per GPU) \\
Number of Epochs              & 4 \\
Sequence Length               & 4096 tokens \\
Tokenizer Pad Token           & Set to \texttt{eos\_token} \\
Precision                     & bfloat16 \\
Loss & Cross-Entropy \\
\bottomrule
\end{tabular}
\caption{Fine-tuning parameter details for \tabqa{}}
\label{fine-tuning}
\end{table}

\subsection{Ablation Study for OTSL Format}

The effectiveness of OTSL for TabVQA is demonstrated through our ablation study (Table \eqref{tab:rep_formats_ablation}). We compare different input formats used to fine-tune as well as infer across both off-the-shelf and fine-tuned LLMs (denoted by the prefix '\tabqa' followed by the format used for finetuning). We show that fine-tuning using OTSL led to the best scores. For each format, we finetune using the same configuration as detailed in Section \eqref{sec:ftn} and report both ANLS and Exact Match (EM) metrics. \tabqa-OTSL shows a gain of around  14 p.p. over \tabqa-HTML on both metrics. 

\begin{table}[htb]
\centering
\begin{tabular}{lccc}
\hline
\textbf{Inferred On} & \textbf{LLM Model \cite{llama3modelcard}} & \multicolumn{1}{c}{\textbf{ANLS}} & {\textbf{EM}}\\ \hline
\multirow{2}{*}{OTSL}       & Off the Shelf  & 25.4	&16.9    \\
                            & \textbf{\tabqa-OTSL}                   & \lightgreenhl{56.5}	& \lightgreenhl{54.0}            \\ \hline
\multirow{2}{*}{HTML}       & Off the Shelf                       &                                   11.0	& 10.3        \\
                            & \tabqa-HTML                   &  42.3	& 39.9                                 \\ \hline
\multirow{2}{*}{Plain Text} & Off the Shelf                       &                                  29.8	& 27.8         \\
                            & \tabqa-Plain             &     52.8	& 50.2  \\ \hline                               
\end{tabular}
\caption{Comparing the impact of fine-tuning on different table representations on the WTQ test set with \lightgreenhl{best} scores highlighted.}
\label{tab:rep_formats_ablation}
\end{table}

\begin{table*}[htb]
\centering
\setlength{\tabcolsep}{2.8pt}
\begin{tabular}{llcccccccc}
\toprule
\multirow{2}{*}{{Type}} & \multirow{2}{*}{{Dataset}} & \multicolumn{2}{c}{{FinTabNet}} & \multicolumn{2}{c}{{PubTabNet}} & \multicolumn{2}{c}{{PubTables}} & 
\multicolumn{2}{c}
{{TORQUE}} \\
\cmidrule(lr){3-4} \cmidrule(lr){5-6} \cmidrule(lr){7-8}
\cmidrule(lr){9-10}
 & & {TEDS-S} & {TEDS} & {TEDS-S} & {TEDS} & {TEDS-S} & {TEDS} & {TEDS-S} & {TEDS} \\
\midrule
\multirow{4}{*}{{End-to-end VLMs}} &  SmolVLM\cite{smolvlm}      & -     & 18.0    & -     & -     & -     & 32.0  & 0.00 & 0.05  \\
& SmolDocling \cite{nassar2025smoldocling}    & 81.0    & 52.0    & -     & -     & {65.0}    & \lightbluehl{88.0}  & 0.05 & 4.50  \\
& Granite Vision \cite{granitevision} & -     & 54.0  & -     & -     & -     & 70.0  & - & -  \\
& MTL-TabNet \cite{ly2023end}     & \lightgreenhl{{98.8}} & -     & \lightgreenhl{{97.9}} & \lightgreenhl{{96.7}} & \lightbluehl{96.7} & \lightgreenhl{96.2} & - & -\\
\midrule
 \multirow{2}{*}{TSR + OCR} & EDD \cite{zhong2020image}            & 90.6  & -     & 89.9  & 88.3  & -     & -     & - & - \\
  & LGPMA \cite{lgpma}            & -  & -     & 96.7  & 94.6  & -     & -     & - & - \\
  \midrule
\multirow{2}{*}{TSR + PDF Parsing} & TableFormer*\cite{tableformer}     & 96.8  & \lightbluehl{89.0}    & 96.7 & {93.6}  & -     & {84.0}   & - & - \\
 & VAST*\cite{huang2023improving}           & \lightbluehl{98.6} & \lightgreenhl{{98.2}} & {97.2} & \lightbluehl{96.0}    & -     & -  & - & -   \\ \midrule
Doubly Decoupled & \textbf{\name (Ours)}   & {98.2} & {55.9}  & \lightbluehl{97.5} & 54.4    & \lightgreenhl{{97.7}} & 54.8 &  \lightgreenhl{{85.4}} & \lightgreenhl{{63.6}}
    \\
\bottomrule
\end{tabular}
\caption{Comparison of TEDS-S and TEDS scores for different table reconstruction methods across popular benchmarks, along with \bench{}. Approaches marked by * use PDF-parsing for content extraction.  The \lightgreenhl{best} and \lightbluehl{second-best} results are highlighted.}
\label{ted_table}
\end{table*}

%% file: sec_R2/5_conclusion.tex
\section{Results and Discussions}
\label{sec:Results}

\textbf{On Table Reconstruction: } We evaluate the effectiveness of our approach, \name{}, across multiple table reconstruction benchmarks. Specifically, \name{} generates an HTML tag sequence, which is subsequently converted into OTSL, capturing both structure and content. Table \eqref{ted_table} presents a comparison of \name{} with state-of-the-art methods, grouped into end-to-end VLMs, conventional decoupled approaches, and our proposed doubly decoupled framework. The evaluation spans FinTabNet, PubTabNet, PubTables, and \bench{} datasets using both TEDS-S and TEDS metrics. Results show that \name{} achieves competitive structural alignment, as reflected in consistently strong TEDS-S scores across all benchmarks, while also highlighting the advantages of our doubly decoupled design. While \name{} outperforms most VLMs in TEDS, it does not surpass MTL-TabNet, which lacks generalizability, or methods like VAST \cite{vast} and TableFormer \cite{tableformer} that exploit PDF parsing for near-perfect text fidelity. Our reliance on OCR introduces the main bottleneck. Although \name{}’s modularity allows easy replacement of the OCR component to boost TEDS scores. As shown in the supplementary material, OCR ablation demonstrates that ground-truth–mapped content achieves near-perfect TEDS, confirming that reconstruction itself is reliable and OCR is the only limitation. Overall, \name{} shows strong structural understanding and achieves solid performance on \bench{}, even with challenging images, while VLM-based methods such as SmolVLM \cite{smolvlm} and SmolDocling \cite{nassar2025smoldocling} remain difficult to extend to multilingual data.

\begin{table}
\centering
\begin{tabular}{lcc}
\hline
\multirow{2}{*}{\textbf{Baseline}} &  \multicolumn{2}{c}{\textbf{Fine-tuned on}}\\ \cmidrule(lr){2-3}
& \textbf{HTML} & \textbf{OTSL} \\

\hline
UDOP \cite{tang2022unifying}  & \lightgreenhl{47.2} & - \\
Pix2struct \cite{lee2022pix2struct} &  39.8 & -\\
DocOwl \cite{hu2024mplug_docowl}  & 26.9 & -\\
Kosmos \cite{peng2023kosmos2}  & 32.4 & -\\
Donut \cite{kim2021ocrfree}&  18.8 & - \\ 
TAPAS \cite{holmgren2012tapas} &  \lightbluehl{46.4} & - \\  
Mistral \cite{mistral2023mistral7b} & - & \lightbluehl{29.9} \\
SOLAR \cite{kim2023solar} & - & 12.3 \\
\textbf{\tabqa{}} & 42.3 & \lightgreenhl{56.5}\\ 
\hline
\end{tabular}
\caption{Comparison of ANLS scores on the WTQ test set. The \lightgreenhl{best} and \lightbluehl{second-best} results are highlighted for clarity.}
\label{qa_tab}
\end{table}
\begin{table}[]
    \begin{center}
        \setlength{\tabcolsep}{1pt}
            \begin{tabular}{llc}
                \toprule
                \textbf{Type} & \textbf{Model} & \textbf{FinTabNetQA} \\ \midrule
                \multirow{7}{*}{Open-source} 

                & BLIP-2~\cite{blip2} & 0.4 \\
                & CogVLM-1k~\cite{wang2023cogvlm} & 4.8 \\
                & CogAgent-VQA~\cite{cogagent} & 22.8 \\
                & SPHINX-v1-1k~\cite{lin2023sphinx} & 3.2 \\
                 & LLaVA-1.5~\cite{liu2023llava-1.5} & 0.8 \\
                & QWEN-VL-Chat~\cite{qwen} & 29.6 \\
                & QWEN-VL~\cite{qwen} & 34.0 \\
                 \cmidrule(lr){1-3}
                \multirow{3}{*}{Closed-source} 
                 & SPHINX-MoE-1k \cite{kim2024tablevqa_bench} & \lightbluehl{36.0} \\
                & SPHINX-v2-1k \cite{kim2024tablevqa_bench} & 31.2 \\
                & SPHINX-MoE \cite{kim2024tablevqa_bench} & 2.8 \\
                \cmidrule(lr){1-3}
                \multirow{2}{*}{Ours} & \textbf{\name{} + \tabqa-HTML} & {29.2} \\
                & \textbf{\name{} + \tabqa-OTSL} & \lightgreenhl{45.2}\\
                \hline
                 \multirow{2}{*}{\textbf{Skyline}} & GT-HTML + TARQA-HTML & \textbf{51.2} \\
                & GT-OTSL + TARQA-OTSL  & \textbf{69.2}\\
                \bottomrule
            \end{tabular}
    \end{center}
    \caption{Comparison of relieved accuracy scores for TabVQA on FinTabNetQA. The results include many baselines alongside our method. Additionally, for skyline, we provide Ground Truth (GT) HTML and OTSL inputs to the corresponding \tabqa{} variant. The \lightgreenhl{best} and \lightbluehl{second-best} scores are highlighted.}
    \label{table:fintabnetqa_scores}
\end{table}

\begin{table*}[ht]
\centering
\begin{tabular}{llcccccc}
\hline
\textbf{Approach} & \textbf{VLM / LLM} & \textbf{Model Parameters} & \textbf{Relieved-Acc.} & \textbf{EM} & \textbf{ANLS} \\ 
\hline


   & Qwen-2.5-VL Instruct \cite{bai2025qwen2} & 7B   & \lightgreenhl{47.40} & \lightgreenhl{46.20} & \lightgreenhl{66.40} \\ 
 \textbf{VLMs} & InternVL-3\_5 \cite{wang2025internvl3} & 8B   & 17.54 & 16.35 & 19.19 \\ 
\textbf{(End-to-End)}  & Paligemma-2 \cite{steiner2024paligemma} & 3B   & 10.20 & 07.10 & 13.90 \\ 
 & SmolVLM-Instruct \cite{smolvlm} & 3B   & 14.45 & 00.95 & 14.45 \\
\hline

\multirow{6}{*}{\textbf{DELTA + LLM}} 
   & Qwen-2.5-Hindi \cite{qwen2.5-14b-hindi} & 14B & 21.30 & 20.40 & 26.50 \\ 
 & Mistral \cite{mistral2023mistral7b} & 7B & 6.64 & 6.16 & 9.00 \\ 
 & HiTQA-mBart \cite{pal2024table} & 611M & 11.22 & 1.43 & 1.91 \\ 
 & HiTQA-M2M \cite{pal2024table} & 484M & 5.73 & 0.24 & 0.24 \\ 
 & mBERT \cite{pires2019multilingual} & 179M & 0.47 & 0.24 & 0.95 \\ 
 & TAPAS \cite{holmgren2012tapas} & 110M & 4.50 & 4.50 & 6.60 \\ \hline
 \multirow{2}{*}{\textbf{Ours}} 
 & \textbf{\name{}(HTML) + TARQA-HTML} & 8B &  12.80  & 12.09 & 16.11 \\
 & \textbf{\name{}(OTSL) + TARQA-OTSL} & 8B &  
 \lightbluehl{27.49} & \lightbluehl{23.93} & \lightbluehl{36.49} \\ 
 
\hline

\multirow{2}{*}{\textbf{Skyline}} 
&  GT HTML + TARQA-HTML & 8B & 28.44  & 28.20 & 31.75 \\  & GT OTSL + TARQA-OTSL & 8B & \textbf{63.51} & \textbf{60.43} & \textbf{76.30}\\
\hline

\end{tabular}
\caption{Comparative results on \bench{} for the TabVQA task. We also report our results with both HTML and OTSL variants of \tabqa{}, along with ground-truth inputs to \tabqa{} as the skyline. The \lightgreenhl{best} and \lightbluehl{second-best} results are highlighted for clarity. }
\label{torque_Results}
\end{table*}

\textbf{On TabQA and TabVQA:} In Table~\eqref{qa_tab}, we evaluate the impact of structured outputs on the downstream TabQA task. Specifically, we compare \tabqa-OTSL and \tabqa-HTML, both of which take DELTA’s corresponding reconstructed outputs. To ensure consistency, we report HTML baseline results from prior work, all fine-tuned on the WTQ dataset. For OTSL baselines, we strictly fine-tune LLMs using the same setup as \tabqa{}. The results clearly show that OTSL-based inputs achieve significantly higher ANLS scores, outperforming the strongest HTML baseline by 9.3 p.p.. We evaluate the performance of our integrated pipeline, \name{} + \tabqa{}, on the FinTabNetQA dataset \cite{kim2024tablevqa_bench} for the TabVQA task. As shown in Table \eqref{table:fintabnetqa_scores}, our approach achieves higher relieved accuracy than all other baselines. We adopt relieved accuracy as the evaluation metric because it accounts for semantically equivalent answers that may differ in surface form (e.g., numerical formatting, currency symbols, or minor textual variations). The skylines (with ground-truth sequences) outperform all other methods. Notably, the OTSL-based skyline achieves the highest score, indicating that more accurate OTSL-based table reconstruction (higher TEDS) directly translates to stronger downstream performance in TabVQA.


\textbf{On Multilingual TabVQA:} To showcase the multilingual capability of our framework, we report results on the \bench{} dataset using different categories of models. Specifically, we evaluate both end-to-end VLMs and several LLMs, where the latter take as input the outputs of \name{} along with the question to be asked. Table~\eqref{torque_Results} presents the results, demonstrating that our approach consistently outperforms all VLMs by more than 10 p.p. except Qwen-2.5 VL Instruct \cite{bai2025qwen2}. Qwen achieves stronger performance only because its pre-training corpus includes Hindi data, which gives it an inherent advantage. For LLMs with Hindi capability, none can surpass our approach. This improvement is notable as \name{} achieves zero-shot performance on \bench{}, with no components trained on Hindi. Furthermore, the skylines (obtained by feeding the ground-truth sequences to \tabqa{}) achieve substantially higher performance than Qwen \cite{bai2025qwen2} and all other VLMs, highlighting that a decoupled module is considerably more effective than end-to-end VLMs. This also indicates that if the inputs to \tabqa{} are of higher quality (i.e., better OCR), as in the case of ground-truth HTML, the downstream performance can be significantly benefited. Qualitative examples for all tasks are included in the supplementary material.

\section{Conclusion}
\label{sec:conclusion}
In summary, we present a comprehensive pipeline for table understanding and reasoning. We begin with \name{}, a doubly decoupled table reconstruction framework that separates structure recognition from OCR and disentangles physical and logical TSR, producing a compact OTSL representation. \name{} achieves high TEDS-S scores and surpasses recent VLM-based approaches. We further introduce a lossless HTML-to-OTSL conversion for interoperability and \tabqa{}, an LLM fine-tuned on OTSL for TabVQA, which achieves strong results on WTQ and FinTabNetQA. Finally, we showcase the performance of \name{} and \name{} + \tabqa{} on the curated Hindi benchmark \bench{}, where our zero-shot approach significantly outperforms other models, including LLMs with inherent Hindi understanding. Collectively, these contributions define a flexible and extensible framework for high-fidelity multilingual table reconstruction and TabVQA.
\label{sec:Conclusion}

\section{Limitations and Future Work}
\label{sec:future}
While \name{} demonstrates good TEDS-S scores, the overall TEDS scores can still be improved. This is due to the current OCR support in the pipeline, which introduces errors that not only affect the TEDS score but can also impact downstream TabVQA performance. Improving OCR quality will push TEDS closer to 100\%. This, in turn, would enable \name{} + \tabqa{} to consistently achieve scores close to skyline performance.  In this way, we will distil the challenges of black-box VLMs into a more well-defined formulation, making Multilingual TabVQA both transparent and debuggable. Our experiments are limited to English and Hindi benchmarks, but the modular design and OCR support will enable extension to other languages. 
Future work will focus on enhancing reasoning for complex queries, such as abstractive QA, thereby broadening the scope and impact of our approach.
\label{sec:Future}

%% file: sec_R2/6_acknowledgement.tex
\section{Acknowledgement}
\label{sec:Acknowledgement}

We acknowledge BharatGen and the Indian Institute of Technology Bombay for providing resources and support for the project. Jahanvi Rajput’s PhD is supported by the Prime Minister’s Research Fellowship (PMRF). 

%% file: sec_R2/supplementary.tex
\clearpage
\setcounter{page}{1}

\twocolumn[
\begin{center}
\Large \textbf{Tables Decoded: \name{} for Structure, \tabqa{} for Understanding \\ Supplementary Material}
\end{center}
\vspace{1em}
]

\label{sec:Supplementary}

\section{Symbols and abbreviations}

Table~\eqref{def_abb} presents the terminology used in the proposed approaches, along with their full forms and the corresponding tasks in which they are applied. Table~\eqref{abb} lists all the abbreviations used throughout the paper, providing readers with an easy reference to understand the terminology better.

\begin{table*}[h!]
\centering
\renewcommand{\arraystretch}{1.3}
\begin{tabular}{>{\bfseries}m{2.5cm} p{7cm} p{5cm}}
\toprule
\textbf{Abbreviation} & \textbf{Description} & \textbf{Task Used For} \\
\midrule
\name{} & \textbf{D}oubly d\textbf{E}coupled tab\textbf{L}e recons\textbf{T}ruction \textbf{A}pproach (a proposed approach) & Table Reconstruction, measured by TEDS-Structure and TEDS Scores \\
\hline
\tabqa{} & \textbf{TA}ble structu\textbf{R}e-aware \textbf{Q}uestion \textbf{A}nswering \newline(a fine-tuned LLM on OTSL sequences) & TabQA \newline Table-based Question Answering \\
\hline
TORQUE & \textbf{T}able \textbf{O}riented \textbf{R}econstruction and \textbf{Q}uestion-answering \textbf{U}pon d\textbf{E}vanagari (curated benchmark)  &  Hindi Table Reconstruction \newline Hindi TabVQA \\
\hline
DELTA + \tabqa{} & DELTA converts Table Image to OTSL \newline \tabqa{} answers Questions with OTSL & Decoupled VQA task \\
\hline
\tabqa{}-OTSL & \tabqa{}
 Fine-tuned on OTSL sequences. & TabVQA \\
 \hline
 \tabqa{}-HTML & \tabqa{}
 Fine-tuned on HTML sequences. & TabVQA \\ 
 \hline
 GT OTSL + \tabqa{}-OTSL & Ground Truth OTSL given to \tabqa-OTSL & TabVQA \\
 \hline
 GT HTML + \tabqa{}-HTML & Ground Truth OTSL given to \tabqa-OTSL & TabVQA \\
 
 \bottomrule
\end{tabular}
\caption{Abbreviations and Their Descriptions}
\label{def_abb}
\end{table*}

\begin{table}
\centering
\begin{tabular}{lc}
\toprule
\textbf{Abbreviation} & \textbf{Description}  \\
\midrule
\multirow{2}{*}{ANLS} & Average Normalized \\ 
& Levenshtein Similarity \\
EM   & Exact Match \\
GT & Ground Truth \\
LLMs & Large Language Models \\
OCR  & Optical Character Recognition \\
\multirow{2}{*}{OTSL} & Optimized Table \\ 
  & Structure Language \\
p.p. & percentage point \\
 \multirow{2}{*}{SPRINT} &Script-agnostic Structure \\
    & Recognition in Tables \\
TabQA & Table Question Answering \\
TabVQA & Table Visual Question Answering \\
TATR & Table Transformer \\
 \multirow{2}{*}{TEDS} &Tree Edit Distance \\ 
   & -based Similarity \\
TSR  & Table Structure Recoginition \\
VLMs & Vision Language Models \\
WTQ  & WikiTableQuestions \\
\bottomrule
\end{tabular}
\caption{General abbreviations used in the paper.}
\label{abb}
\end{table}

\section{Motivation for Doubly Decoupled Approach}
Table Reconstruction approaches can be divided into: 
\begin{itemize}
    \item \textbf{End-to-end VLMs} In conventional table understanding pipelines, physical structure, logical structure, and content recognition (OCR) are often bundled together into a monolithic framework. We refer to this as the VLMs paradigm, where all three aspects: cell boundaries, spanning relations, and textual extraction are jointly modeled. While such end-to-end systems simplify design, they are typically hard to interpret, debug, and adapt across domains and languages.

    \item \textbf{Conventional Decoupling:} Recent approaches such as CascadeTabNet \cite{cascadetabnet}, TATR \cite{tatr-pub-1m}, GTE \cite{gte}, TableFormer \cite{tableformer}, and LGPMA \cite{lgpma} attempt to decouple structure recognition from content recognition. However, within the structure recognition stage, the physical structure (rows, columns, cell boundaries) and the logical structure (spanning cells, header associations, merged cells) remain entangled. This coupling often limits flexibility and makes it challenging to localise errors, as mispredictions in geometry and semantics influence each other.

    \item \textbf{Our Doubly-Decoupled Framework: (\name{})} We introduce a finer decomposition by separating physical and logical structure recognition into independent stages. In the structure recognition stage, the physical structure (cell boundaries) and the logical structure (cell arrangements) should be loosely coupled. The physical structure is inherently tied to image coordinates, whereas the logical structure only requires predicting the arrangement and relationships among cells. By decoupling the two, we can keep the logical layer language-agnostic, while isolating coordinate-dependent errors in the physical layer. This separation improves flexibility, simplifies debugging, and allows each component to be strengthened independently. Specifically, TATR models the physical layout, while SPRINT captures logical relations. Their outputs are then combined into a complete table structure, which is subsequently passed to OCR for content extraction. This double decoupling offers greater control and modularity.
\end{itemize}

\section{\name{} in detail}
\label{param:delta}
Our proposed framework, \name{}, tackles the long-standing challenge of disentangling physical and logical structures in table structure recognition. Unlike traditional approaches, where geometric layout (rows, columns, cell boundaries) and logical layout (cell arrangement, spans, header associations) are tightly coupled, \name{} predicts them independently and then combines their outputs in a principled manner. Specifically, we employ SPRINT for logical structure prediction, which generates compact OTSL/HTML sequences and offers an ideal balance of speed, accuracy, and language independence, making it robust across multilingual and noisy documents. For the physical structure, we use TATR, a DETR-based model pre-trained on large-scale datasets, that leverages only row and column predictions to ensure clean grid alignment. These two components are integrated to reconstruct complete HTML tables with explicit bounding boxes, row spans, and column spans, making the system modular, interpretable, and extensible. Beyond structure prediction, we also examine the role of OCR for content recognition and conduct comprehensive ablations to validate and justify our design choices.

\subsection{SPRINT for Logical Structure}

\begin{algorithm}[]
\caption{Converting OTSL Matrix (logical structure) and cell boxes (physical structure) into HTML Sequence.}
\label{alg:sprinttatr}
\begin{algorithmic}[1]
\State \textbf{Input:} OTSL matrix $M$ of size $R \times C$, list of cell bounding boxes $Cells$
\State \textbf{Output:} HTML table string $H$, list of structured cells $S$
\State $H \gets$ ``\texttt{<table><tbody>}'' 
\For{$i = 1 \to R$}
    \State Append ``\texttt{<tr>}'' to $H$
    \For{$j = 1 \to C$}
        \If{$M[i,j] = \texttt{C}$}
            \State $cell \gets Cells[i,j]$
            \State $(rs, cs) \gets \texttt{get\_cell\_spans}(M, i, j)$
            \If{$rs > 0 \wedge cs > 0$}
                \State Extend $td$ to cover row and column spans
                \State Append \texttt{<td rowspan=rs+1 colspan=cs+1 bbox=cell>} to $H$
            \ElsIf{$rs > 0$}
                \State Extend $td$ vertically
                \State Append \texttt{<td rowspan=rs+1 bbox=cell>} to $H$
            \ElsIf{$cs > 0$}
                \State Extend $td$ horizontally
                \State Append \texttt{<td colspan=cs+1 bbox=cell>} to $H$
            \Else
                \State Append \texttt{<td bbox=cell>} to $H$
            \EndIf
        \ElsIf{$M[i,j] = \texttt{N}$}
            \State Append ``\texttt{</tr>}'' to $H$
        \EndIf
    \EndFor
\EndFor
\State Append ``\texttt{</tbody></table>}'' to $H$
\State \Return $H$
\end{algorithmic}
\end{algorithm}

SPRINT is an image-to-sequence model that employs a Global Context Attention (GCA)-based encoder and a transformer-based decoder to generate compact sequence representations (OTSL/HTML) of logical table structures. We adopt SPRINT for logical structure prediction because it achieves an ideal balance of speed, accuracy, and language independence. Unlike methods that rely heavily on OCR or language-specific cues, SPRINT focuses solely on the structural layout, making it inherently robust across multilingual and noisy document settings. This design aligns perfectly with our objective of decoupling physical and logical structure, as SPRINT cleanly predicts the cell arrangement without being confounded by text semantics. Moreover, since it has already demonstrated state-of-the-art performance on table structure recognition benchmarks, it provides a reliable backbone for our framework. We report ablations reported on SPRINT to demonstrate these strengths in Table \eqref{tab:ablation}. For further implementation details, we refer the reader to the original SPRINT \cite{kudale2025sprint} paper.

\begin{table*}[]
\centering
\begin{tabular}{|c|c|c|c|c|c|}
\hline
\textbf{Test} &
  \textbf{Training} &
  \textbf{SPRINT Config} &
  \textbf{\begin{tabular}[c]{@{}c@{}}TEDS-S \\ Simple\end{tabular}} &
  \textbf{\begin{tabular}[c]{@{}c@{}}TEDS-S \\ Complex\end{tabular}} &
  \textbf{\begin{tabular}[c]{@{}c@{}}TEDS-S \\ Overall\end{tabular}} \\ \hline
\multirow{4}{*}{PubTabNet}    & PubTabNet    & *Layers: 3, Shape: 32*128  & 97.91          & 91.17          & 94.61          \\ \cline{2-6} 
                              & PubTabNet    & Layers: 4, Shape: 32*128  & 98.12          & 92.84          & 95.53          \\ \cline{2-6} 
                              & All          & Layers: 6, Shape: 32*128  & \textbf{98.11} & 92.98          & 95.60          \\ \cline{2-6} 
                              & All          & Layers: 6, Shape: 128*128 & 98.00          & \textbf{93.32} & \textbf{95.71} \\ \hline
\multirow{4}{*}{FinTabNet}    & FinTabNet    & Layers: 6, Shape: 32*32   & \textbf{98.39} & 94.57          & 96.41          \\ \cline{2-6} 
                              & FinTabNet    & Layers: 6, Shape: 32*128  & 98.30          & 97.46          & 97.88          \\ \cline{2-6} 
                              & All          & Layers: 6, Shape: 128*128 & 98.31          & 97.73          & 98.01          \\ \cline{2-6} 
                              & FinTabNet    & Layers: 6, Shape: 128*128 & 98.35          & \textbf{97.74} & \textbf{98.03} \\ \hline
\multirow{4}{*}{PubTables-1M} & PubTables-1M & Layers: 6, Shape: 32*128  & 98.19          & 92.69          & 95.50          \\ \cline{2-6} 
                              & All          & Layers: 8, Shape: 32*128  & 98.88          & 93.34          & 96.00          \\ \cline{2-6} 
                              & All          & Layers: 6, Shape: 32*128  & 98.87          & 94.80          & 96.75          \\ \cline{2-6} 
                              & All          & Layers: 6, Shape: 128*128 & \textbf{98.92} & \textbf{96.54} & \textbf{97.68} \\ \hline
\end{tabular}
\caption{Results on different test sets for the SPRINT component of \name{} trained on various datasets. The training set of 'All' refers to the combined training dataset of all three datasets. The config is dictated by two parameters, mainly the number of decoder layers and the shape (dimensions) of the input image, which is resized in the preprocessing stage. * indicates that the maximum permissible length of prediction was set to 192 for that experiment, and the length was set to 224 otherwise. All the results are reported on the canonical validation set of PubTabNet \cite{otsl} and canonical test sets of FinTabNet \cite{otsl} and PubTables-1M \cite{otsl}}
\label{tab:ablation}
\end{table*}

\subsection{TATR for Physical Structure}
To extract the physical structure, we use the TATR~\cite{tatr-pub-1m} V1.1 model pre-trained on FinTabNet, PubTabNet, and PubTables-1M. TATR, built on DETR~\cite{detr}, predicts six classes, of which we only leverage \texttt{table-row} and \texttt{table-column} to estimate the rows and columns. For inference, we set the detection threshold to 0.25 and apply non-maximum suppression (NMS) with an IoU threshold of 0.25 on \texttt{table-row} predictions to minimize overlap and improve consistency. The resulting values are then aligned with the output sequence predicted by SPRINT, ensuring coherence between physical and logical structures.

\subsection{TATR and SPRINT for Complete TSR}
\label{sec:TATR and SPRINT for Complete TSR}
This step is responsible for integrating the logical structure (tag sequence predicted by SPRINT) with the physical structure (list of bounding boxes corresponding to detected rows and columns) to produce a final HTML representation of the table. Each \texttt{<td>}  element in the output is annotated with its bounding box coordinates, as well as rowspan and colspan attributes whenever merged cells are detected. While row and column bounding boxes intersect to form candidate cells, the crucial constraint is that there exists a one-to-one mapping between each logical cell predicted by SPRINT and its corresponding physical bounding box. The algorithm enforces this alignment to guarantee that both spatial positioning and spanning attributes are preserved.

Formally, as seen in Algorithm \eqref{alg:sprinttatr}, it takes as input an OTSL matrix $M$ (generated by SPRINT, note that it can be converted to HTML in a lossless manner, but since SPRINT directly gives an OTSL string, we leverage that initially) and a set of bounding boxes $Cells$ (derived from TATR). The OTSL matrix encodes the table layout, where each entry specifies whether a position corresponds to a cell (\texttt{C}), a new row marker (\texttt{N}), or is empty. The process begins by initializing an empty HTML string $H$. For each entry $(i,j)$ in $M$:

\begin{itemize}
    \item If it corresponds to a cell, the bounding box $cell$ is retrieved from $Cells$. The function \texttt{get\_cell\_spans} computes the extent of row and column spans by checking consecutive overlaps in the SPRINT output.

    \item If spans are present, the bounding box is extended accordingly, and an HTML $td$ tag with the appropriate rowspan and/or colspan attributes is generated and appended to $H$.

    \item If no spans are present, a simple $td$ tag with its bounding box is appended. In both cases, the processed bounding box is stored in the $bbox$ attribute.
\end{itemize}

Whenever an entry is marked as \texttt{N}, the algorithm closes the current row and begins a new one. After all entries are processed, the final HTML string is completed.

The output consists of the structured HTML string $H$, which encodes the table with explicit row and column spans that maintain the bounding boxes associated with each logical cell. This design ensures that even complex tables with merged rows/columns are reconstructed faithfully, while retaining both logical order (from SPRINT) and spatial grounding (from TATR).

\subsection{OCR Ablations}
The TSR module of \name{} achieves highly reliable structure predictions, with TEDS-Structure scores exceeding 95\% on standard datasets. However, the overall TEDS score is hampered by OCR quality, making OCR the primary performance bottleneck. OCR errors arise from noise, complex layouts, slanted text, and special symbols. Since OCR follows TSR in our pipeline, its choice is critical. To assess this impact, we compare two widely used engines: Tesseract \cite{tesseract} and EasyOCR \cite{easyocr}. Tesseract provides broad multilingual support, modularity, and ease of integration but suffers from lower accuracy on noisy data, lacks GPU acceleration, and is slow at inference. In contrast, EasyOCR is GPU-compatible, nearly three times faster, and consistently more accurate. It integrates seamlessly with detection and structure recognition modules, offers greater control over outputs, and is modular enough to be replaced with CRNN-based or fine-tuned models. Empirically, EasyOCR achieves consistently stronger results (Table~\ref{ocr_table}), improving TEDS scores of almost all the datasets. The weighted average increases by 12.45 p.p. across FinTabNet, PubTabNet, FinTabNetQA, and \bench, which also translates into higher TabVQA accuracy. The slight drop in performance on PubTabNet stems from its relatively clean images with few OCR errors, rather than a limitation of EasyOCR. In this setting, Tesseract and EasyOCR perform nearly identically, with negligible differences in TEDS scores. In contrast, PubTables-1M poses a greater challenge: its large scale makes Tesseract impractically slow for experiments, whereas EasyOCR strikes a balance between speed and accuracy, achieving a reasonable TEDS score of 54.8. Table~\eqref{tab:lat} highlights the latency comparison, underscoring EasyOCR’s superior efficiency. Accordingly, EasyOCR is used as the default OCR module in our \name{} pipeline, though it can be readily replaced with a stronger alternative. 

To isolate OCR as the primary performance bottleneck, we prepared Ground Truth (GT) mapped predictions, where the content of each predicted HTML cell was replaced with the corresponding ground-truth content while preserving the predicted structure. Care was taken to ensure accurate row-wise fidelity and cell mapping. This setup is equivalent to DELTA predictions followed by perfect OCR. As shown by the resulting TEDS scores (Table \ref{ocr_table}) above 90\% in most cases. This shows that the content errors are almost entirely attributable to OCR. This study clearly indicates that integrating a stronger OCR module can substantially boost the table reconstruction quality of \name, effectively narrowing the problem.

\begin{table}[h!]
\centering
\begin{tabular}{lccc}
\toprule
\textbf{Dataset} & \textbf{Tesseract} & \textbf{Easy OCR} & \textbf{GT Mapped}\\
\midrule
FinTabNet   & 41.5 & \textbf{55.9} & 91.2\\
PubTabNet   & \textbf{54.4} & 53.0 &  87.8 \\
FinTabNetQA & 70.0 & \textbf{84.0} &  92.5 \\
PubTables-1M & - & 54.8 & 86.8 \\
TORQUE      & 40.8 & \textbf{63.6} & 83.5\\
\bottomrule
\end{tabular}
\caption{TEDS scores across different OCR modules and ground truth content mapped to the structure of \name{}. Results show consistent improvements on FinTabNet, FinTabNetQA, and TORQUE datasets.}
\label{ocr_table}
\end{table}

\begin{table}[]
    \centering
    \begin{tabular}{lc}
    \hline
        \textbf{OCR Engine}  & \textbf{Average Latency}   \\
        & \textbf{per image (in secs)}\\
        \toprule
        Tesseract & 13.9 \\
        EasyOCR & \textbf{4.3 }\\
        \bottomrule
    \end{tabular}
    \caption{Average Latency comparison of Tesseract and EasyOCR per table image in seconds calculated on the FinTabNet test set.}
    \label{tab:lat}
\end{table}




\begin{table*}[ht]
\centering
\renewcommand{\arraystretch}{1.2}
\resizebox{\textwidth}{!}{%
\begin{tabular}{lp{11cm}}
\toprule
\textbf{Metric} & \textbf{Definition} \\
\midrule
\textbf{Tree Edit Distance-based Similarity (TEDS)} & Measures the similarity between predicted and ground truth HTML table structures using tree edit distance. It evaluates both the structure and content of tables, which we use to evaluate \name{} pipeline. \\
\textbf{TEDS-Structure (TEDS-S)} & A TEDS variant that evaluates only the HTML tag sequence, ignoring content. We use it to assess \name{}'s table structure recognition. \\
\textbf{Average Normalized Levenshtein Similarity (ANLS)} & Measures textual similarity between predicted and ground truth answers using the normalised Levenshtein distance, providing partial credit for near matches. We use this metric to evaluate \tabqa{} on the WTQ dataset. \\
\textbf{Exact Match (EM)} & A strict binary metric that returns 1 if the predicted answer matches the ground truth exactly, and 0 otherwise. \\
\textbf{Relieved-Accuracy} & This metric considers predictions correct if any normalised form matches the ground truth, ignoring units and formatting. It emphasizes semantic equivalence while evaluating \name{} + \tabqa{} on the FinTabNetQA dataset for the TabVQA task. \\
\bottomrule
\end{tabular}
}
\caption{Overview of metrics used for evaluating \name{} and \tabqa.}
\label{tab:metric_definitions}
\end{table*}

\section{Evaluation Metrics}

Table~\eqref{tab:metric_definitions} lists the metrics used for evaluation. TEDS and TEDS-S are applied to the table reconstruction task on PubTabNet, PubTables-1M, FinTabNet, and TORQUE; TEDS-S captures structural fidelity, while TEDS score takes into account both structure and content, making them well-suited for measuring both layout accuracy and content alignment. For the TabQA and TabVQA tasks on WTQ, FinTabNetQA, and TORQUE, we report ANLS, EM, and Relieved Accuracy, all of which range from 0 to 100. These metrics are standard in QA benchmarks, reflecting exact correctness (EM), tolerance to minor variations (ANLS), and robustness to different semantic answer formats (Relieved Accuracy). Together, they ensure fair and comprehensive comparisons across diverse methods.

\section{OTSL Ablation}

\begin{figure*}[h]
    \centering
    \begin{subfigure}{\linewidth}
        \centering
        \includegraphics[width=\linewidth]{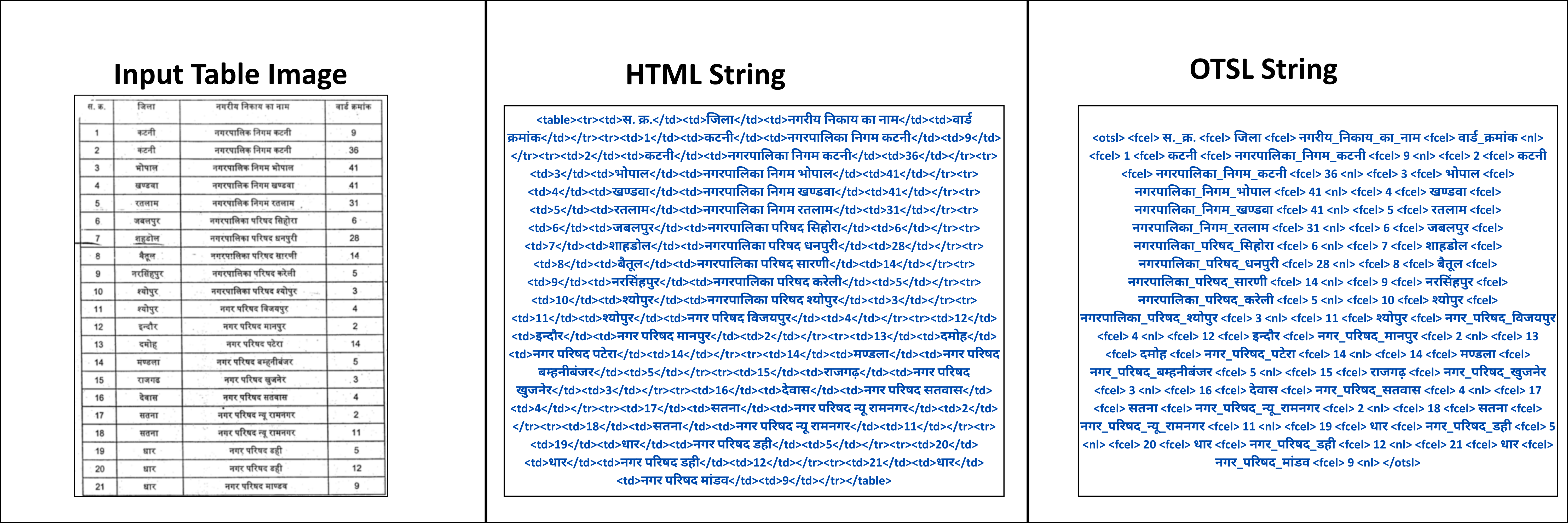}
        \caption{HTML Character Count: 438, OTSL Character Count: 409}
        \label{otsl_subfig_1}
    \end{subfigure}
    \hfill
    \begin{subfigure}{\linewidth}
        \centering
        \includegraphics[width=\linewidth]{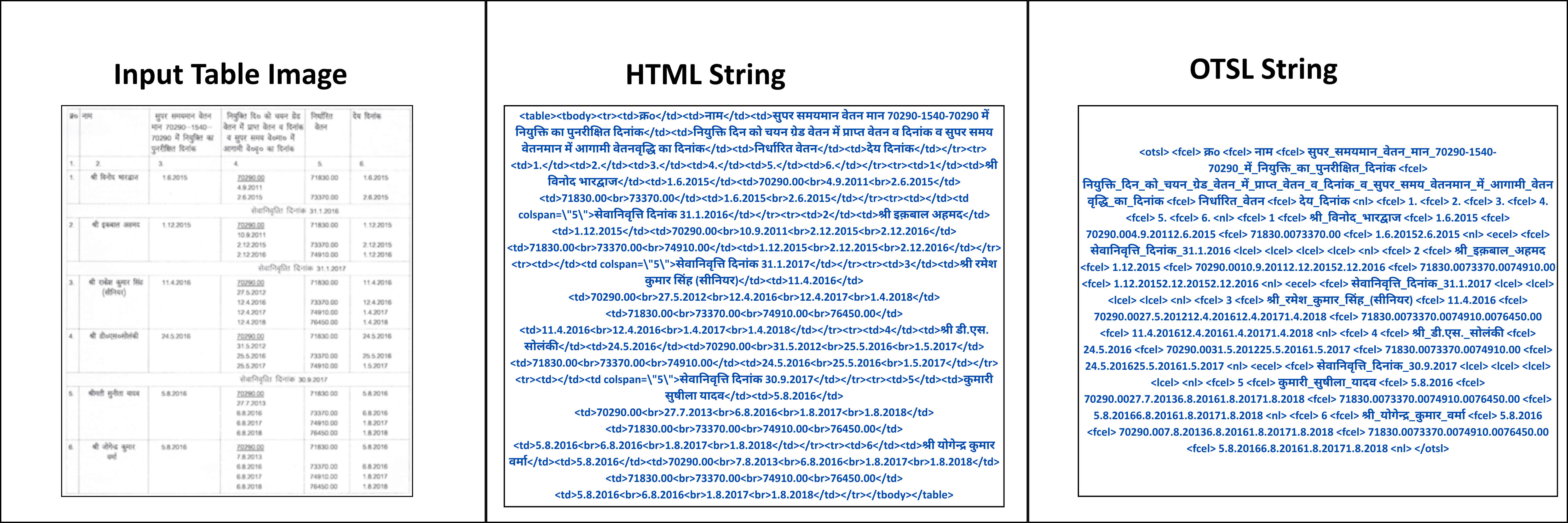}
        \caption{HTML Character Count: 1607, OTSL Character Count: 1430}
        \label{otsl_subfig_2}
    \end{subfigure}
    \caption{Qualitative examples from the TORQUE dataset illustrating the compactness of the OTSL format in terms of Character Counts from the input image.}
    \label{fig:otsl_tor}
\end{figure*}

Figure \eqref{fig:otsl_tor} presents qualitative examples illustrating the compactness of the OTSL format relative to HTML. We include two scenarios: one featuring a large table and another involving a large table with a more complex structure. These examples demonstrate that OTSL not only preserves structural fidelity but also provides a significantly more compact representation than HTML. The compactness of OTSL further benefits LLMs by reducing context length, enabling more efficient and accurate table understanding. In our analysis, several input tables that were incorrectly processed in HTML format were correctly interpreted when encoded in OTSL, highlighting its effectiveness. This reduced representation allows larger tables to be processed without exceeding context limits, ultimately contributing to improved overall accuracy. Finally, the handling of mathematical expressions depends on the OCR module applied after structure recognition; therefore, OTSL itself does not impose any inherent limitations on the extraction of mathematical equations.

\section{Qualitative Results}

In this section, we present qualitative results for table reconstruction on FinTabNet and TORQUE datasets, as well as for TabQA and TabVQA tasks.

\subsection{Table Reconstruction}

\begin{figure*}[h]
    \centering
    \begin{subfigure}{\linewidth}
        \centering
        \includegraphics[width=\linewidth]{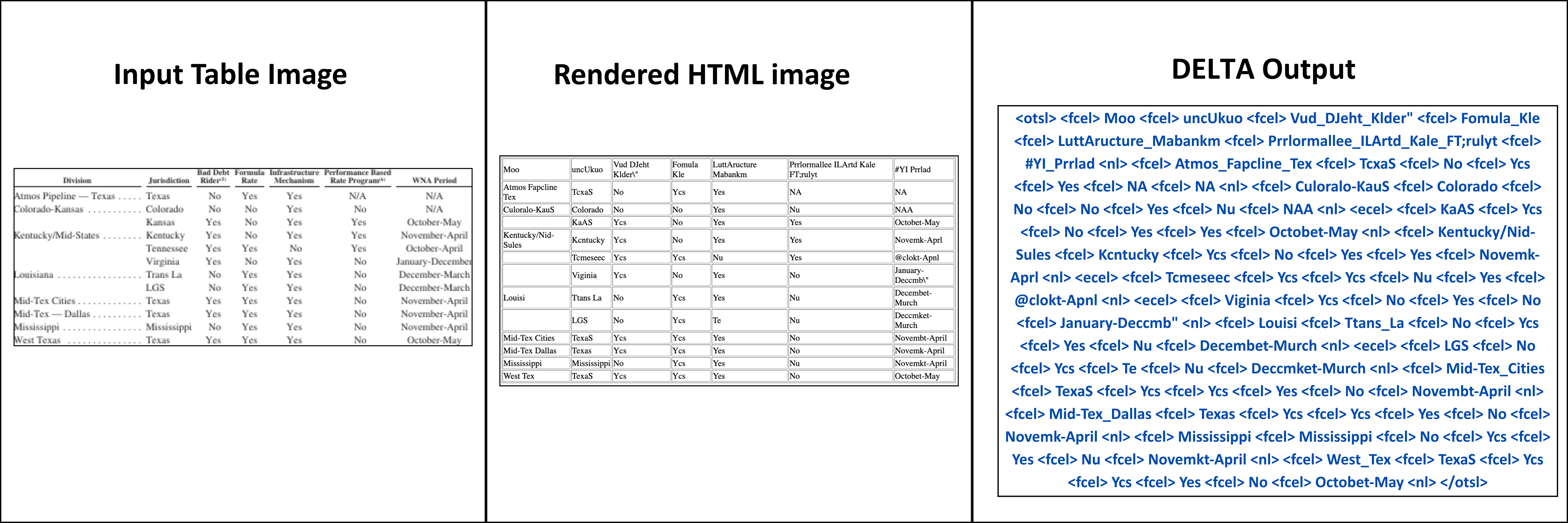}
        \caption{TEDS-S = 100 and TEDS = 80.98}
        \label{tr_subfig_1}
    \end{subfigure}
    \hfill
    \begin{subfigure}{\linewidth}
        \centering
        \includegraphics[width=\linewidth]{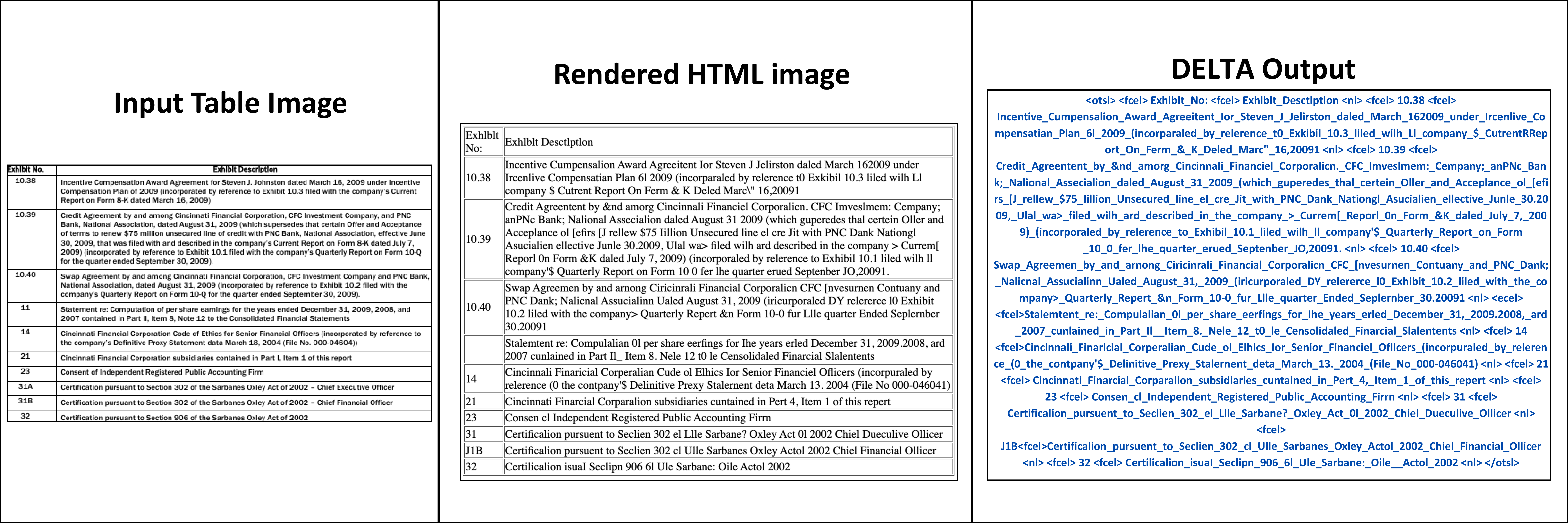}
        \caption{TEDS-S = 100 and TEDS = 88.11.}
        \label{tr_subfig_2}
    \end{subfigure}
    \caption{Qualitative examples from the FinTabNet dataset illustrating table reconstruction outputs produced by \name{} framework.}
    \label{fig:tr_fintabnet}
\end{figure*}

\begin{figure*}[h]
    \centering
    \begin{subfigure}{\linewidth}
        \centering
        \includegraphics[width=\linewidth]{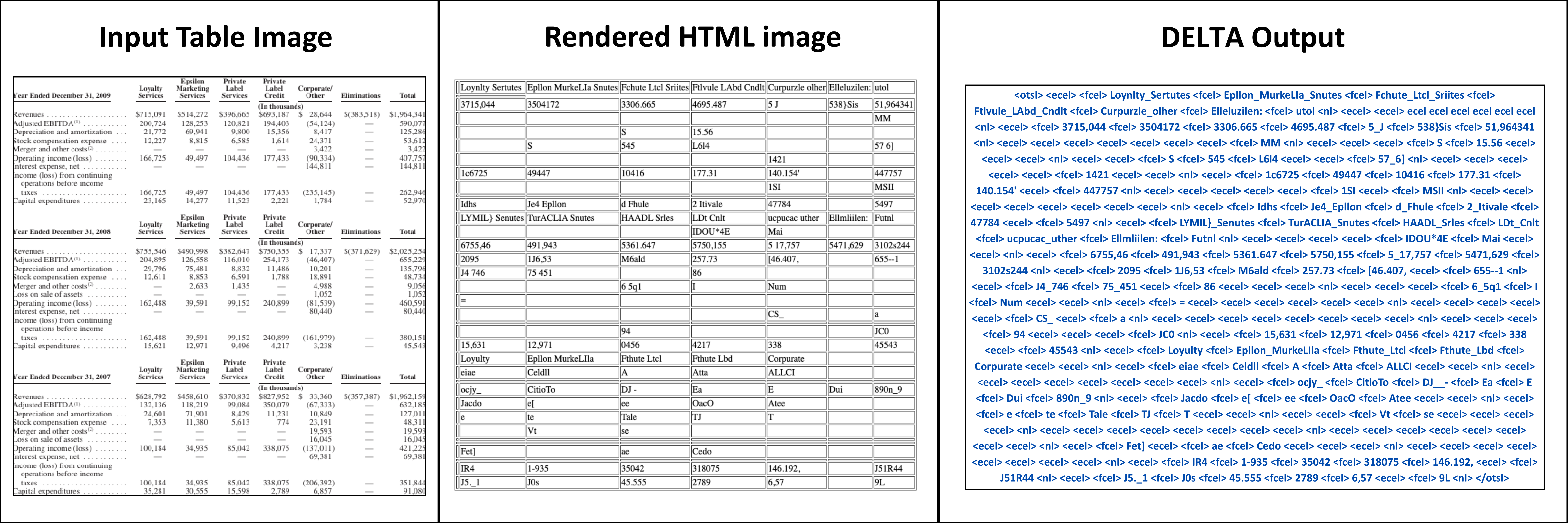}
        \caption{TEDS-S = 95.15 and TEDS = 19.35}
        \label{tr_subfig_3}
    \end{subfigure}
    \hfill
    \begin{subfigure}{\linewidth}
        \centering
        \includegraphics[width=\linewidth]{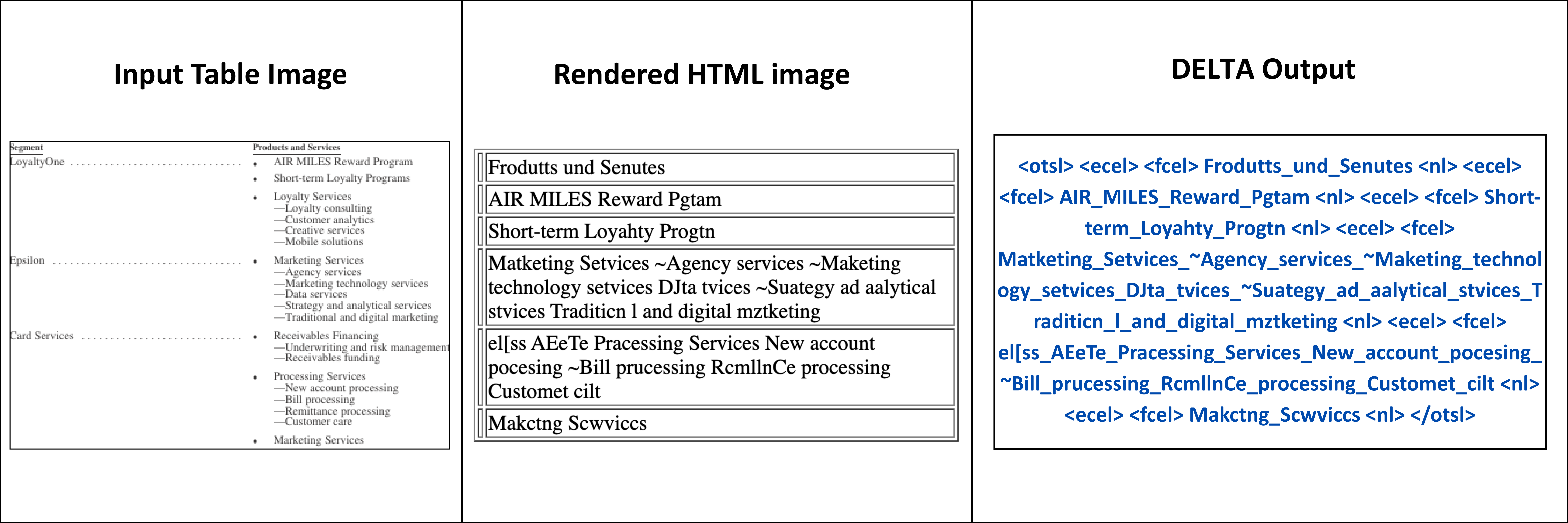}
        \caption{TEDS-S = 26.09 and TEDS = 19.95}
        \label{tr_subfig_4}
    \end{subfigure}
    \caption{Inaccurate qualitative examples from the FinTabNet dataset illustrating table reconstruction outputs generated by \name{}.}
    \label{fig:tr_fintabnet}
\end{figure*}

Figure~\eqref{tr_subfig_1} demonstrates a successful case of \name{}, where the table image has a clean layout with multiple columns and well-separated numeric content. This results in perfect structural accuracy (TEDS-S = 100) and a high overall score (TEDS = 80.98). Similarly, Figure~\eqref{tr_subfig_2} shows another positive example, with a clear two-column structure and well-defined row–column divisions, yielding TEDS-S = 100 and TEDS = 88.11. In contrast, Figure~\eqref{tr_subfig_3} highlights a failure case arising from OCR extraction errors: while the structural score remains high (TEDS-S = 95.15), the overall content fidelity is poor (TEDS = 19.35). Figure~\eqref{tr_subfig_4} presents another challenging example, where low image resolution and a borderless table with bullet points hinder accurate reconstruction, leading to low scores (TEDS-S = 26.09, TEDS = 19.95).

\begin{figure*}[h]
    \centering
    \begin{subfigure}{\linewidth}
        \centering
        \includegraphics[width=\linewidth]{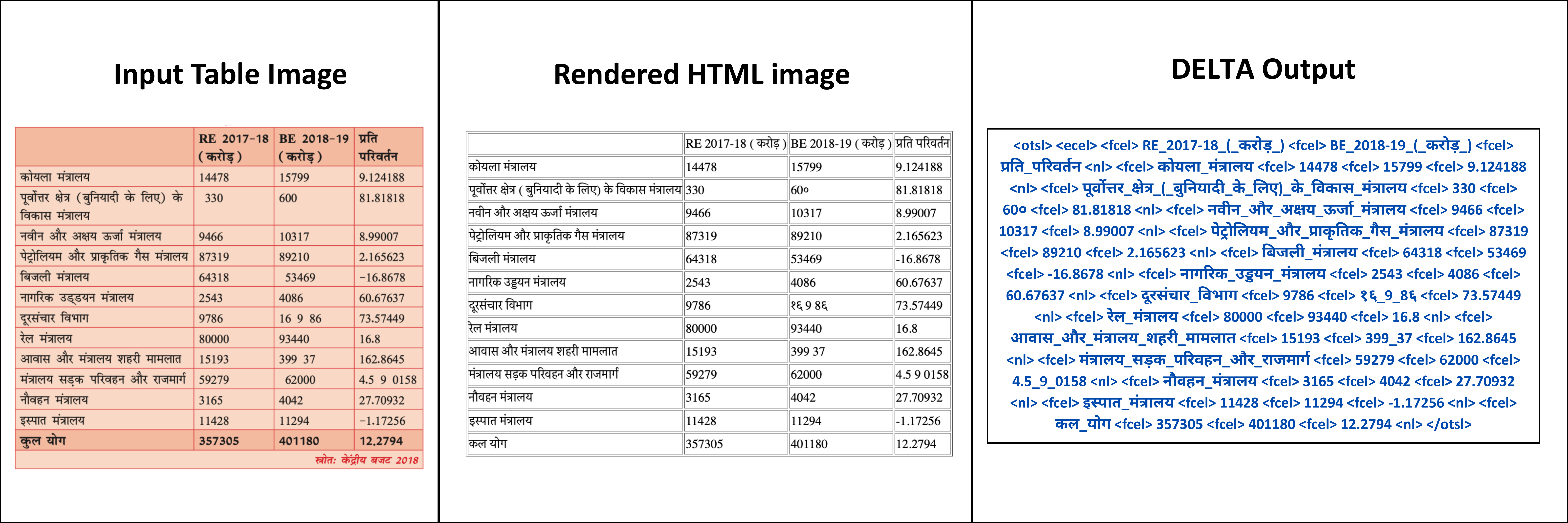}
        \caption{Shows a prediction that closely aligns with the input table, achieving a high TEDS-S and TEDS score of 97.22 and 96.06, respectively.}
        \label{subfig1}
    \end{subfigure}
    \hfill
    \begin{subfigure}{\linewidth}
        \centering
        \includegraphics[width=\linewidth]{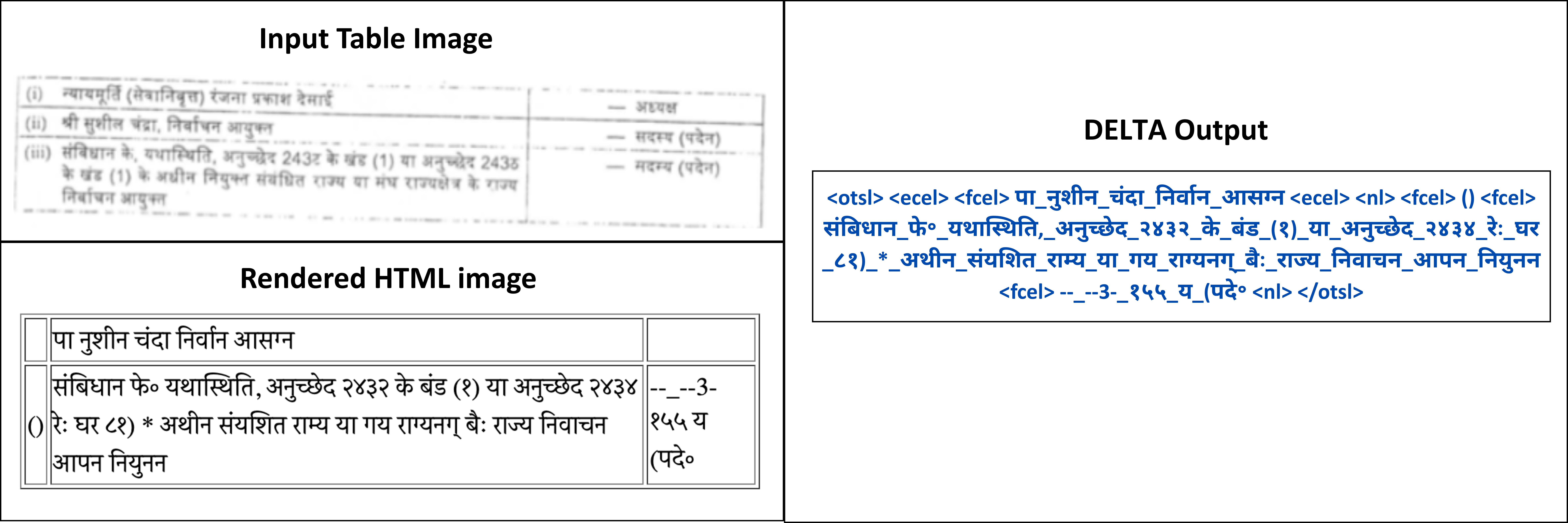}
        \caption{Depicts a challenging case with slanted text and low-resolution input, leading to notable differences between the predicted and input tables and a lower TEDS-S and TEDS score of 44.44 and 14.60, respectively.}
        \label{subfig2}
    \end{subfigure}
    \caption{Qualitative results of the proposed \name{} framework on a TORQUE sample for the table reconstruction task.}
    \label{fig:torque}
\end{figure*}

Similarly, Figure~\eqref{fig:torque} presents qualitative examples of table reconstruction results on the \bench{} dataset. Figure~\eqref{subfig1} demonstrates a successful reconstruction, reflected in a high TEDS-S and TEDS score of 97.22 and 96.06, respectively. In contrast, Figure~\eqref{subfig2} highlights a failure case where slanted text and low-resolution input hinder accurate prediction, resulting in a significantly lower TEDS-S and TEDS score of 44.44 and 14.60, respectively.

\subsection{Table Question Answering}

\begin{figure*}[h]
    \centering
    \begin{subfigure}{\linewidth}
        \centering
        \includegraphics[width=\linewidth]{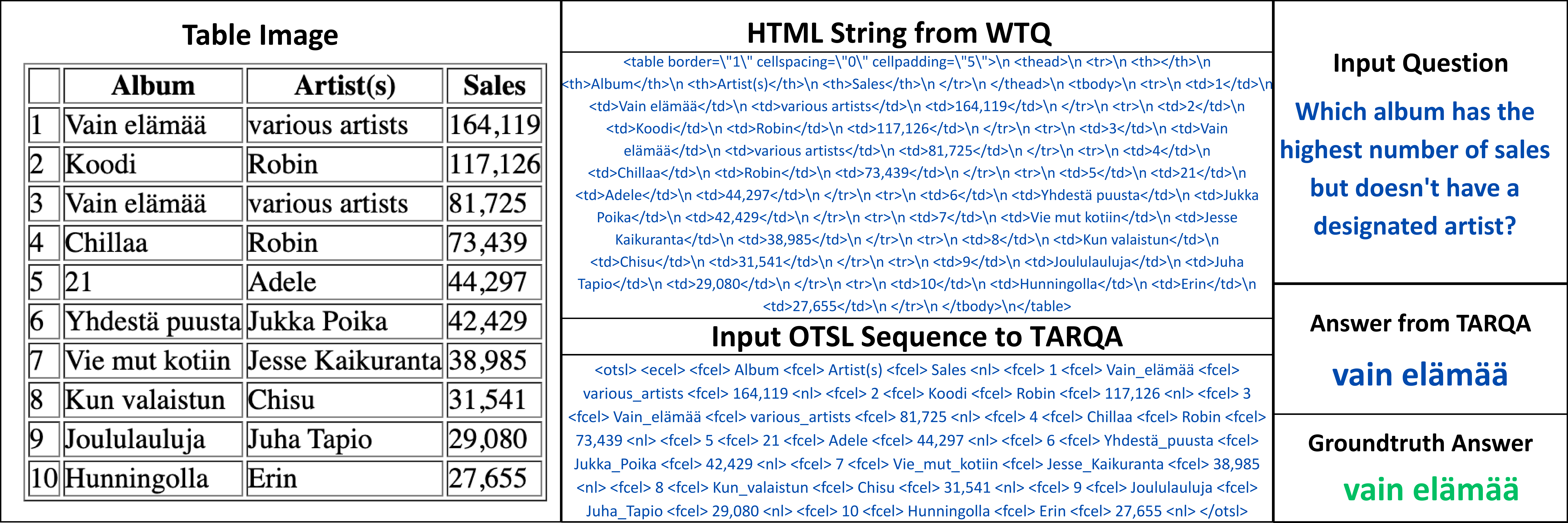}
        \caption{The predicted answers align with the ground truth}
        \label{fig:subfig_1}
    \end{subfigure}
    \hfill
    \begin{subfigure}{\linewidth}
        \centering
        \includegraphics[width=\linewidth]{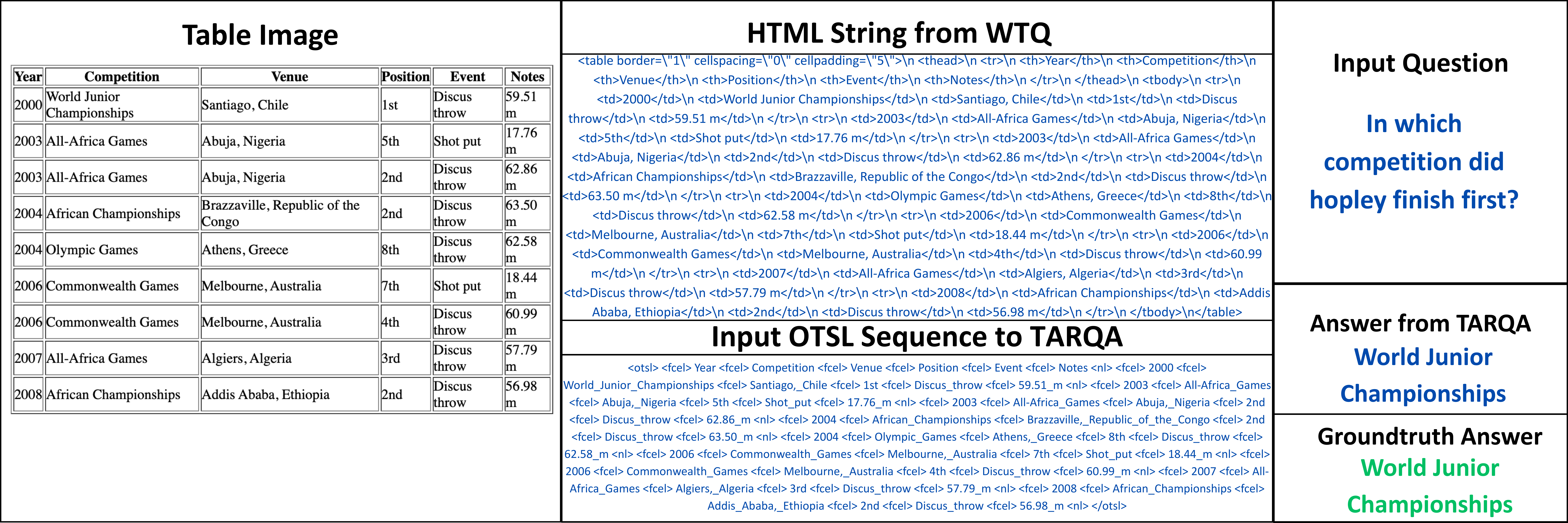}
        \caption{The predicted answers align with the ground truth}
        \label{fig:subfig_2}
    \end{subfigure}
    \caption{Qualitative examples from the WTQ dataset for TabQA tasks, illustrating answers generated \tabqa{-OTSL}.}
    \label{fig:tavqa}
\end{figure*}
\begin{figure*}[h]
    \centering
    \begin{subfigure}{\linewidth}
        \centering
        \includegraphics[width=\linewidth]{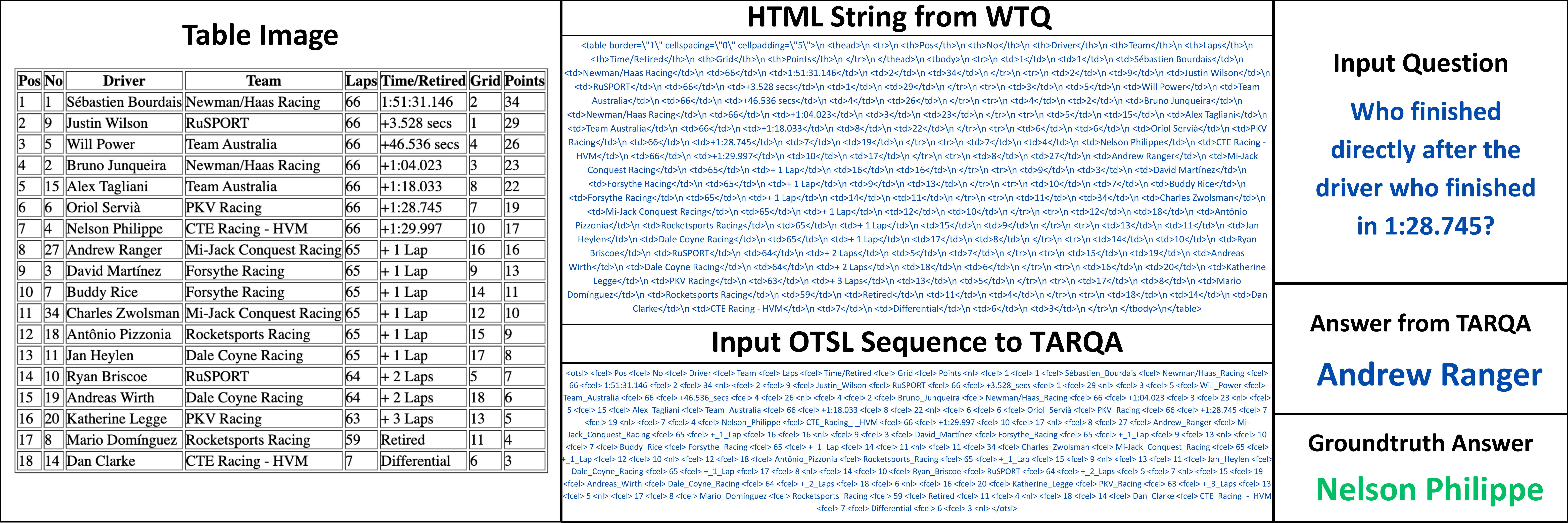}
        \caption{The predictions differ from the ground truth}
        \label{fig:subfig_3}
    \end{subfigure}
    \hfill
    \begin{subfigure}{\linewidth}
        \centering
        \includegraphics[width=\linewidth]{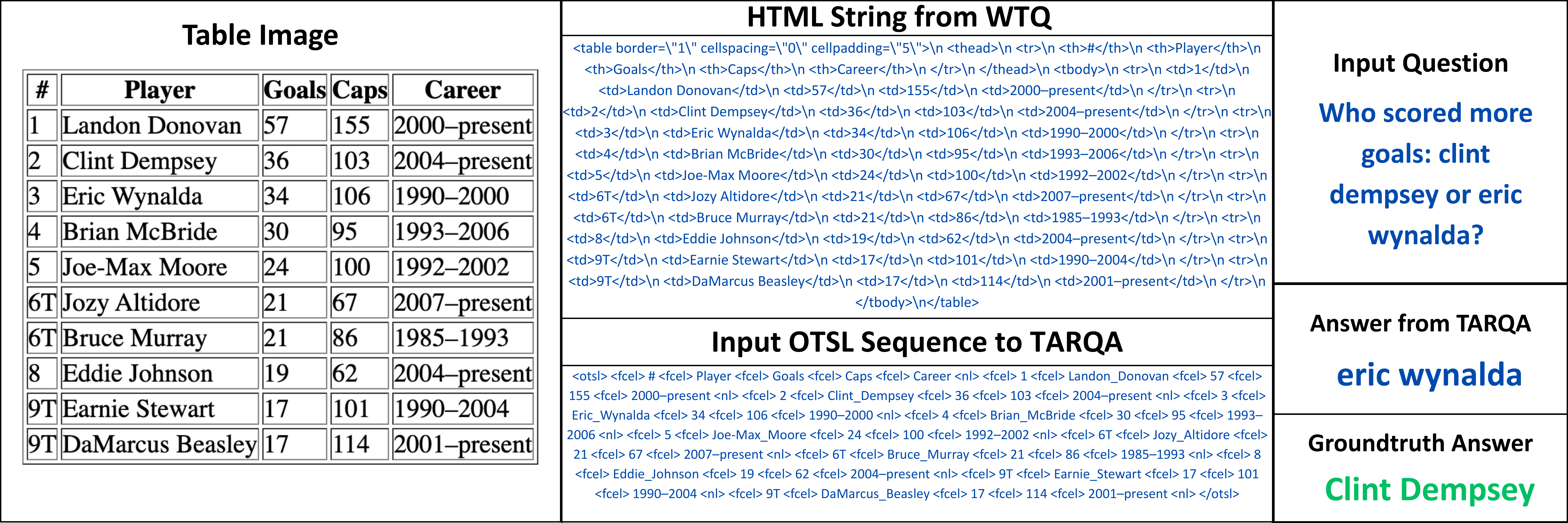}
        \caption{The predictions differ from the ground truth}
        \label{fig:subfig_4}
    \end{subfigure}
    \caption{Qualitative examples from the WTQ dataset for TabQA tasks, illustrating answers generated by \tabqa{-OTSL}.}
    \label{fig:tavqa}
\end{figure*}

Figure~\eqref{fig:subfig_1} and Figure~\eqref{fig:subfig_2} present qualitative examples from the WTQ dataset for TabQA tasks, where the answers predicted by the proposed \tabqa{} framework are consistent with the ground truth. 
In contrast, Figure~\eqref{fig:subfig_3} and Figure~\eqref{fig:subfig_4} illustrate cases where the predictions deviate from the ground truth. For Figure~\eqref{fig:subfig_3}, the model outputs a value occurring immediately after the correct answer, indicating a limitation in its contextual understanding. For Figure~\eqref{fig:subfig_4}, the prediction is incorrect because the question is of a comparative type, for which the model has not been explicitly fine-tuned. \\

\subsection{Table Visual Question Answering}

\begin{figure*}[h]
    \centering
    \begin{subfigure}{\linewidth}
        \centering
        \includegraphics[width=\linewidth]{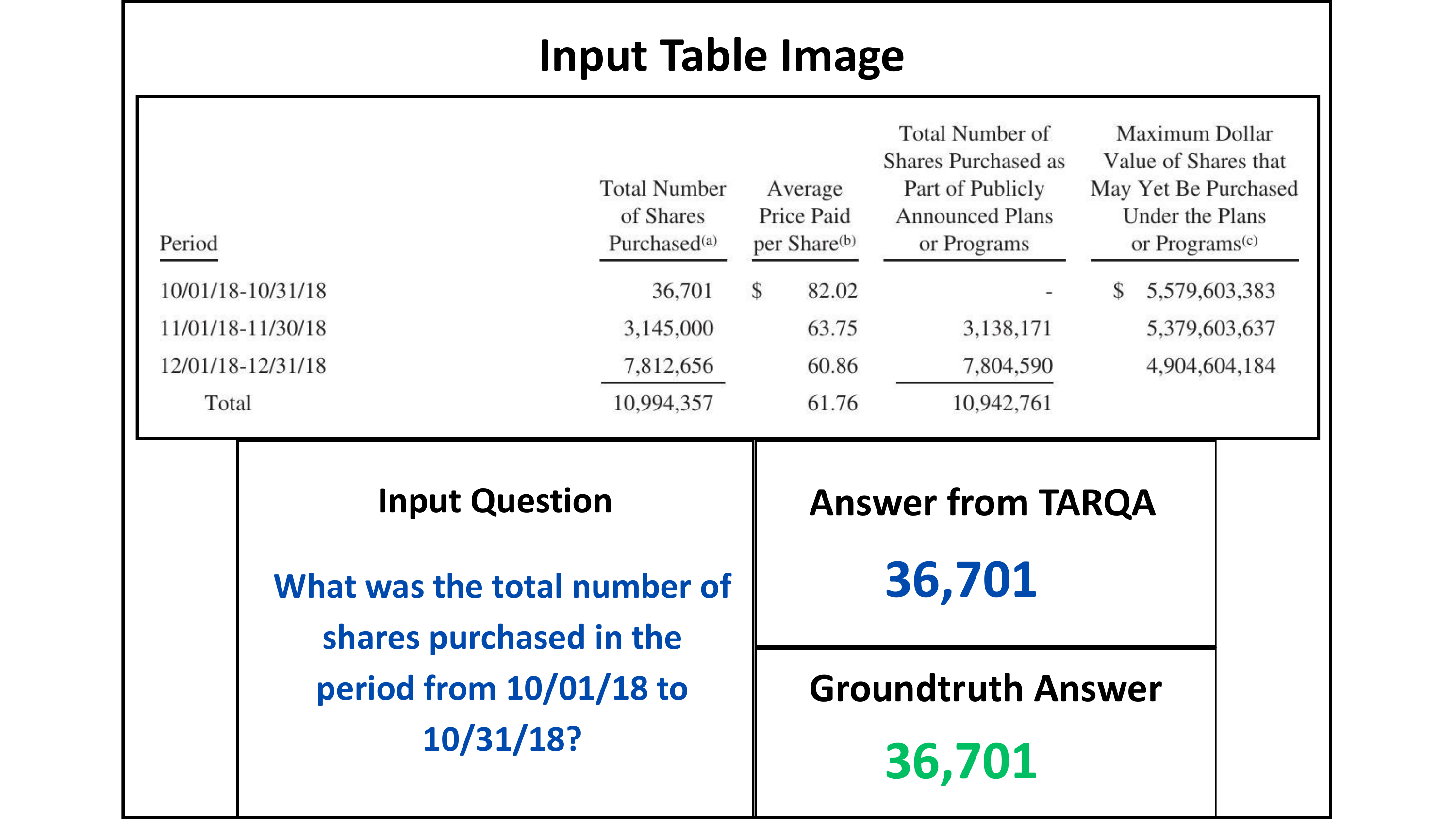}
        \caption{The predicted answers coincide with the ground truth}
        \label{fig:subfig1}
    \end{subfigure}
    \hfill
    \begin{subfigure}{\linewidth}
        \centering
        \includegraphics[width=\linewidth]{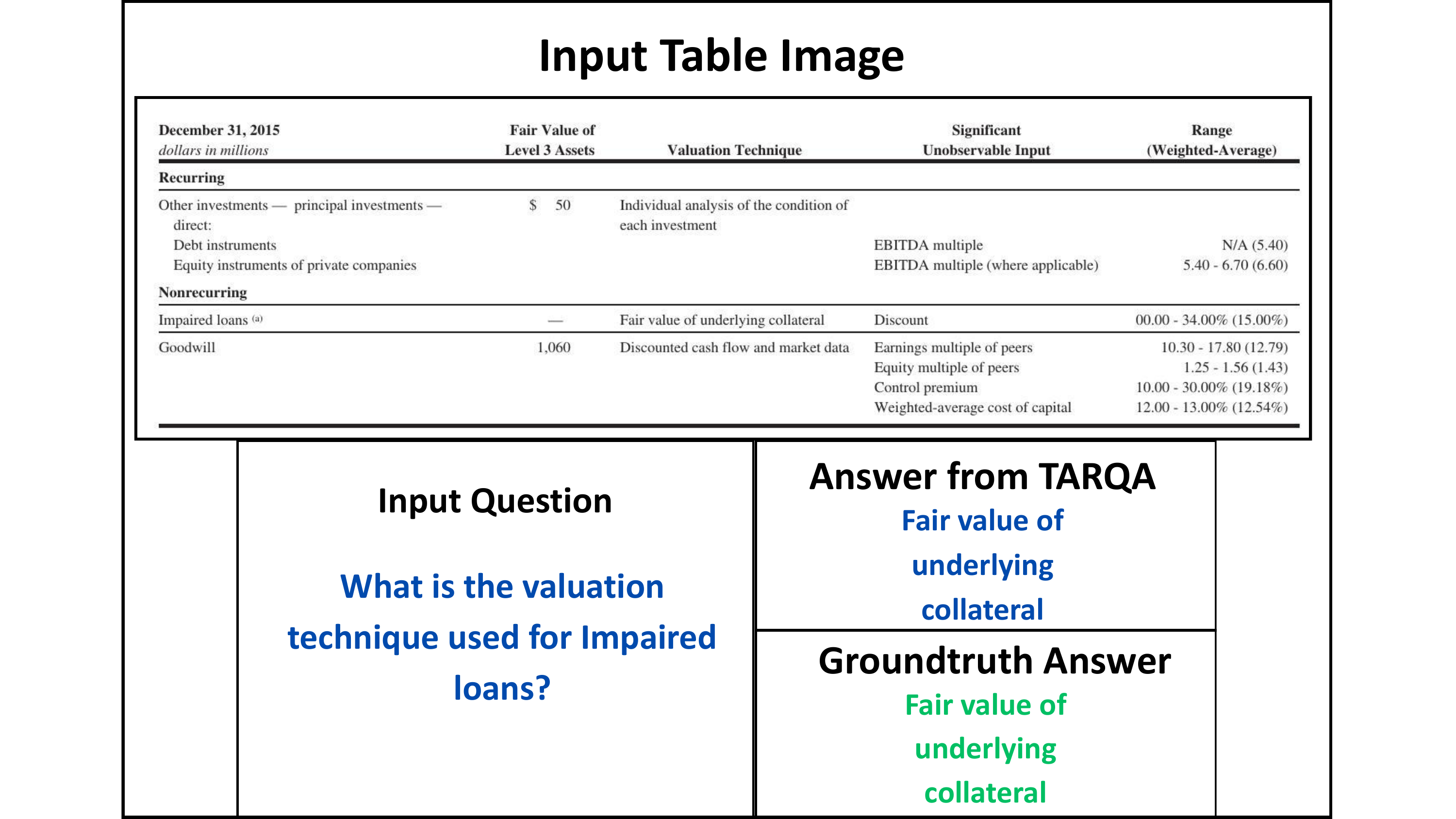}
        \caption{The predicted answers coincide with the ground truth}
        \label{fig:subfig2}
    \end{subfigure}
    \caption{Illustrative TabVQA examples from the FintabnetQA dataset showcasing the performance of \name{}+\tabqa{-OTSL}.}
    \label{fig:tabvqa}
\end{figure*}

\begin{figure*}[h]
    \centering
    \begin{subfigure}{\linewidth}
        \centering
        \includegraphics[width=\linewidth]{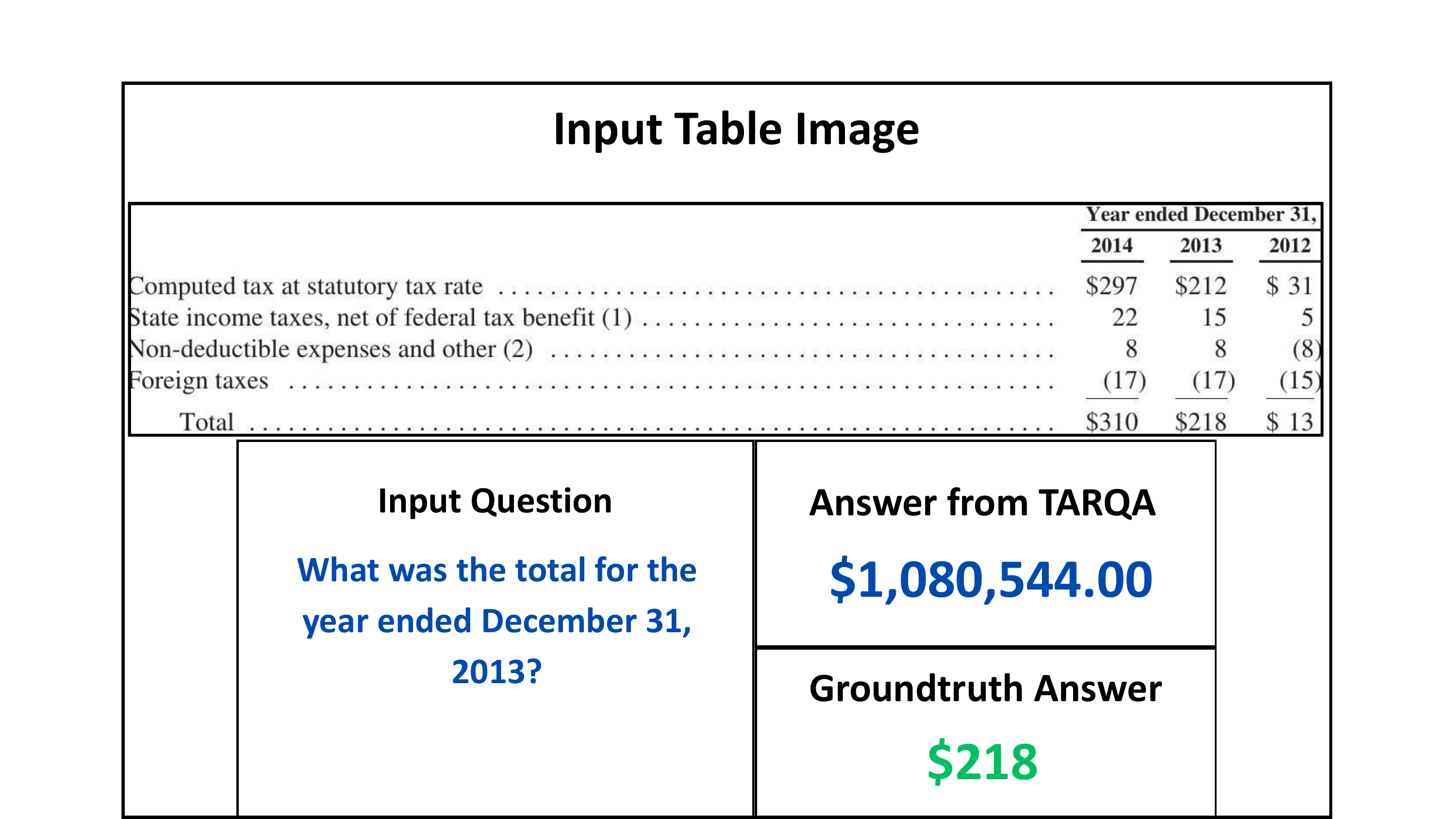}
        \caption{The ground truth and predicted answers are diverging from each other.}
        \label{fig:subfig3}
    \end{subfigure}
    \hfill
    \begin{subfigure}{\linewidth}
        \centering
        \includegraphics[width=\linewidth]{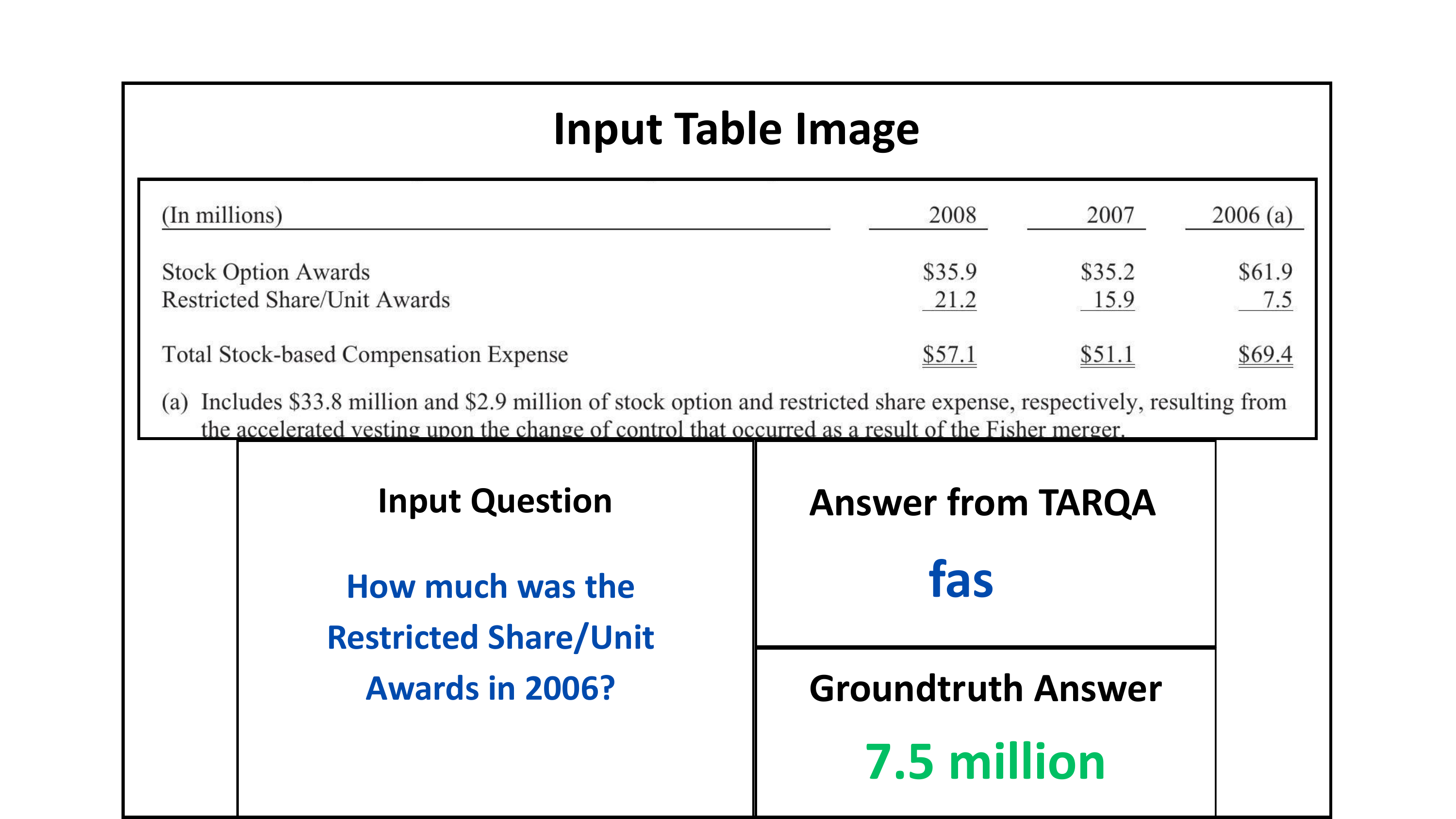}
        \caption{The ground truth and predicted answers are diverging from each other.}
        \label{fig:subfig4}
    \end{subfigure}
    \caption{Illustrative TabVQA examples from the FintabnetQA dataset showcasing the performance of \name{}+\tabqa{-OTSL}.}
    \label{fig:tabvqa}
\end{figure*}

Figure~\eqref{fig:subfig1} and Figure~\eqref{fig:subfig2} show illustrative TabVQA examples from the FintabnetQA dataset using the \name{}+\tabqa{} pipeline. These examples highlight that even with difficult, borderless tables, the framework can accurately predict both numerical and textual answers, aligning with the ground truth. On the other hand, Figure~\eqref{fig:subfig3} and Figure~\eqref{fig:subfig4} showcase instances of divergence. In Figure~\eqref{fig:subfig3}, the model becomes confused due to the complexity of the question type, while in Figure~\eqref{fig:subfig4}, poor image quality leads to incorrect answer. These cases highlight both the strengths and current limitations of the approach, while also indicating clear directions, improved OCR, and enhanced reasoning for future enhancements.

